%% file: arxiv.tex
\pdfoutput=1

\documentclass{article}

\usepackage[preprint]{neurips_2026}

\input{preamble.tex}

\author{%
  Mufan Qiu \\
  University of North Carolina\\at Chapel Hill
  \And
  Genhui Zheng \\
  The University of Texas\\at Austin
  \And
  Yinuo Xu \\
  University of\\Pennsylvania
  \And
  Ruichen Zhang \\
  University of North Carolina\\at Chapel Hill
  \AND
  Ying Ding \\
  The University of Texas\\at Austin
  \And
  Qi Long \\
  University of\\Pennsylvania
  \And
  Tianlong Chen \\
  University of North Carolina\\at Chapel Hill
}

\begin{document}

\maketitle

\input{body.tex}


\newpage
\input{checklist.tex}

\end{document}

%% file: preamble.tex
\usepackage[utf8]{inputenc} 
\usepackage[T1]{fontenc}    
\usepackage{hyperref}       
\usepackage{url}            
\usepackage{booktabs}       
\usepackage{amsfonts}       
\usepackage{nicefrac}       
\usepackage{microtype}      
\usepackage{xcolor}         
\usepackage{amsmath}        
\usepackage{amssymb}        
\usepackage{enumitem}       
\usepackage{graphicx}       
\usepackage{wrapfig}        
\usepackage{array}          

\title{Chreode: A Cell World Model for One\mbox{-}Step Temporal Dynamics and Perturbation Prediction}

%% file: body.tex
\begin{abstract}
Predicting how a cell will change its transcriptional state under a developmental signal or a genetic perturbation is the computational core of in\mbox{-}silico biology and the AI Virtual Cell program. 
Existing approaches either fit static control\mbox{-}to\mbox{-}treated maps that discard time, or solve multi\mbox{-}step ODE / Schr\"odinger\mbox{-}bridge problems on each dataset independently. 
We introduce \textbf{Chreode}, a one-step cell world model that predicts action-conditioned cell-state transitions through a structured residual transition operator.
It shifts distributional evolution from inference time to training time, enabling single-pass generation while preserving a Waddington-inspired decomposition into downhill landscape flow, rotational in-tangent dynamics, and stochastic spread.
The model is pretrained with a shared scVI encoder and a DiT-based dynamics backbone on a 2.4M\mbox{-}cell mouse embryonic atlas spanning $7$ datasets. 
As a fine-tuning initialization, \textbf{Chreode} improves per-target Sinkhorn distance on Weinreb hematopoiesis and Veres islet differentiation over matched scratch models, PI-SDE, and PRESCIENT. 
As a transferable gene-state embedding for GEARS, the pretrained dynamics representation reduces shared-vocabulary DE20 mean squared error on Norman Perturb-seq from $0.2121$ to $0.1858$, a $12.4\%$ relative improvement, without changing the GEARS training procedure. 
We interpret this transfer to perturbation prediction as evidence that pretrained developmental\mbox{-}trajectory dynamics encode differentiation primitives transferable to CRISPR\mbox{-}induced state shifts, since both involve cell\mbox{-}state transitions in a shared latent geometry. 
The pretrained backbone additionally produces zero\mbox{-}shot clonal fate scores on Weinreb that are competitive with strong dynamic\mbox{-}OT baselines.

\end{abstract}

\section{Introduction}
\label{sec:intro}

A modern in\mbox{-}silico perturbation screen needs to query on the order of $10^{4}$ cells, $10^{3}$ drugs or genetic perturbations, and $10$ time points, of order $10^{8}$ (state, action, time) tuples per screen. Existing generative cell\mbox{-}dynamics models cost 40 to 100 ODE or stochastic\mbox{-}bridge steps per query, putting full screens at $10^{10}$ network forward passes. Predicting how a cell will change its transcriptional state given an action and an elapsed time is the computational core of the \emph{AI Virtual Cell} program \citep{bunne2024aivirtualcell}; reaching screen scale requires amortizing the transition itself, not just running the same multi\mbox{-}step solver faster.


We call a model that directly predicts this transition a \emph{cell world model}, by analogy to the state\mbox{-}action\mbox{-}transition world models in reinforcement learning \citep{ha2018worldmodels,hafner2020dreamer}. Given a latent transcriptional state $z_t$ at time $t$, an intervention $a$, and an elapsed time $\Delta$, such a model predicts the distribution of possible future states $z_{t+\Delta}$, equivalently $p(z_{t+\Delta}\mid z_t,\mathrm{do}(a))$. We use the analogy as a predictive\mbox{-}model framing only and do not claim planning or rollout.

Four properties of cell dynamics make a naive application of existing generative recipes fail. (i) \emph{Temporal information is systematically underused}: destructive scRNA\mbox{-}seq yields unpaired population snapshots, yet most perturbation predictors collapse these into static control\mbox{-}to\mbox{-}treated maps and discard the rate of change biology provides for free \citep{lotfollahi2019scgen,lotfollahi2023cpa,roohani2024gears,bunne2023cellot}. (ii) \emph{State and action are not interchangeable}: a drug is an \emph{intervention} in the sense of Pearl's do\mbox{-}calculus \citep{pearl2009causality}, not a latent context to concatenate; simple additive composition \citep{lotfollahi2019scgen,lotfollahi2023cpa} is accurate near attractors but cannot resolve compositional or trajectory\mbox{-}dependent perturbations. (iii) \emph{Inference cost does not amortize with screen scale}: per\mbox{-}query multi\mbox{-}step integration multiplies linearly with the number of (cell, action, time) tuples, so a $10^{2}$\mbox{-}step solver makes a $10^{8}$\mbox{-}query screen cost $10^{10}$ forward passes regardless of per\mbox{-}step efficiency. (iv) \emph{The inductive bias is biological rather than Euclidean}: Waddington's landscape \citep{waddington1957}, cell fate as a ball rolling through a potential with rotational currents, motivates a residual decomposed into a potential gradient, an antisymmetric flow, and a stochastic spread; yet existing flow\mbox{-}matching, Schr\"odinger\mbox{-}bridge, and OT methods \citep{lipman2023flow,tong2023cfm,tang2025branchsbm,cellflow2025} parameterize the velocity field as a generic neural network without this architectural reflection.

\begin{wrapfigure}[19]{r}{0.36\textwidth}
  \vspace{-0.6em}
  \centering
  \includegraphics[width=0.36\textwidth]{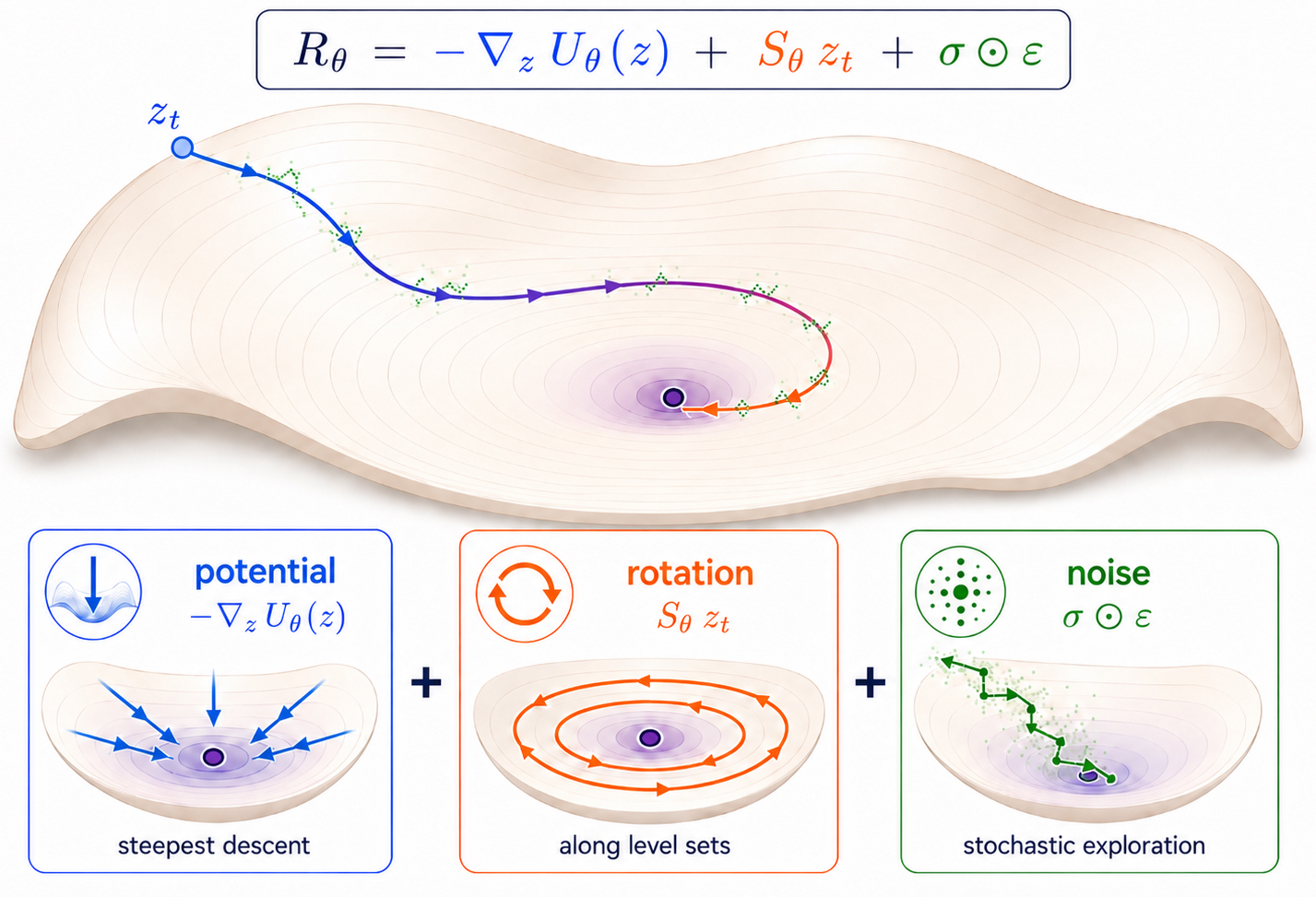}
  \caption{\small The Waddington residual at the heart of Chreode. A latent state $z_t$ traverses a learned landscape via a potential gradient $-\nabla_{z}U_{\theta}$ (blue), an antisymmetric in\mbox{-}tangent rotation $S_{\theta}z_{t}$ (orange), and stochastic spread $\sigma\odot\epsilon$ (green). The full residual is applied as a single one\mbox{-}step update $\hat z_{t+\Delta} = z_t + \alpha(\Delta)R_\theta$, not by ODE rollout.}
  \label{fig:teaser}
  \vspace{-0.8em}
\end{wrapfigure}

Prior work addresses at most two of these properties. Flow\mbox{-}matching and Schr\"odinger\mbox{-}bridge methods \citep{lipman2023flow,tong2023cfm,tang2025branchsbm,cellflow2025} use time, but require multi\mbox{-}step integration and typically lack explicit action conditioning. Perturbation predictors such as scGen, CPA, GEARS, and CellOT \citep{lotfollahi2019scgen,lotfollahi2023cpa,roohani2024gears,bunne2023cellot} model actions, but fit static control\mbox{-}to\mbox{-}treated maps and ignore temporal structure. Cell foundation models such as scGPT, Geneformer, and CellStream \citep{cui2024scgpt,theodoris2023geneformer,sha2025cellstream} learn transferable representations, but do not natively model a transition operator. Concurrent cellular world models, AlphaCell \citep{alphacell2025} and X\mbox{-}Cell \citep{xcell2026}, scale the framing, but rely on multi\mbox{-}step OT\mbox{-}CFM or iterative diffusion denoising. We instead build on the recent \emph{Drifting Models} framework \citep{deng2026drifting}, which shows that a single\mbox{-}forward\mbox{-}pass network can match multi\mbox{-}step diffusion / flow methods by evolving the pushforward distribution \emph{at training time} via a population\mbox{-}level drifting field; we combine this training\mbox{-}time pushforward with a Waddington biological prior and snapshot\mbox{-}atlas pretraining.

\paragraph{Our approach.}
We introduce \textbf{Chreode}\footnote{A \emph{chreode} \citep{waddington1957} is a canalized developmental trajectory through the epigenetic landscape. We use this term because Chreode predicts a single\mbox{-}step transition along a learned canalized flow.}. For an input latent state $z_t$, elapsed time $\Delta$, intervention $a$, and random noise vector $\epsilon$, Chreode predicts a future latent state $\hat z_{t+\Delta}$ by adding a learned residual to the input state:
\begin{equation}
\hat z_{t+\Delta} \;=\; z_t \;+\; \alpha(\Delta)\cdot R_\theta(z_t, \Delta, a, \epsilon),
\qquad
R_\theta \;=\; -\nabla_z U_\theta \;+\; S_\theta\,z_t \;+\; \sigma_\theta\odot\epsilon,
\label{eq:model}
\end{equation}
Here, $R_\theta$ is the learned residual transition, and $\theta$ is the trainable neural network parameters. The first term, $-\nabla_z U_\theta$, is the negative gradient of a learned potential function $U_\theta$ with respect to the latent state $z_t$; it represents downhill motion on a Waddington\mbox{-}style landscape. 
The second term, $S_\theta z_t$, uses a learned antisymmetric operator $S_\theta$ to model rotational or cyclic dynamics. 
In practice, we parameterize this operator as $S_\theta = P_\theta Q_\theta^\top - Q_\theta P_\theta^\top$, where $P_\theta$ and $Q_\theta$ are neural network outputs whose difference of outer products guarantees antisymmetry. 
The third term, $\sigma_\theta\odot\epsilon$, models stochastic population spread, where $\sigma_\theta$ is a learned scale vector, $\epsilon$ is sampled noise, and $\odot$ denotes elementwise multiplication. 
Finally, $\alpha(\Delta)=1-e^{-\Delta/\tau_0}$ is a time gate with learnable time constant $\tau_0$; this gate ensures that the predicted change vanishes when no time has elapsed, because $\alpha(0)=0$.
We pretrain Chreode in two stages: a shared scVI \citep{lopez2018scvi} encoder over a fixed cross\mbox{-}species ortholog vocabulary, then a DiT\mbox{-}based dynamics backbone \citep{peebles2023dit} trained on a 2.4M\mbox{-}cell mouse embryonic atlas of seven datasets and 88 timepoints, using a population\mbox{-}level drifting\mbox{-}field loss \citep{deng2026drifting}.
This loss provides anti\mbox{-}collapse gradients beyond maximum mean discrepancy, denoted MMD, and Sinkhorn Wasserstein\mbox{-}2 distance, denoted $W_2$.


After pretraining, the frozen Chreode backbone provides two distinct transfer modes. As a \emph{fine\mbox{-}tuning initialization} for population time\mbox{-}transition prediction (\S\ref{sec:downstream:weinreb}, \S\ref{sec:downstream:veres}), the pretrained backbone improves per\mbox{-}target Sinkhorn $W_2$ on Weinreb hematopoiesis ($d4$, $d6$) and Veres islet differentiation ($t1$ through $t7$) over matched scratch and over PI\mbox{-}SDE / PRESCIENT \citep{yeo2021prescient} trained on the same shared scVI\mbox{-}128 latent. As a \emph{gene\mbox{-}state embedding} injected into the existing GEARS predictor \citep{roohani2024gears} on Norman Perturb\mbox{-}seq (\S\ref{sec:downstream:norman}), it reduces shared\mbox{-}vocabulary DE20 MSE from $0.2121$ to $0.1858$ ($12.4\%$ relative) without modifying the GEARS training procedure; we interpret this as evidence that the dynamics backbone, pretrained on mouse embryonic differentiation manifolds, encodes primitives that also describe CRISPR\mbox{-}induced state shifts. The pretrained backbone additionally produces zero\mbox{-}shot clonal fate scores on Weinreb (\S\ref{sec:downstream:fate}) competitive with strong dynamic\mbox{-}OT baselines (moscot, WOT, scDiffEq).

\paragraph{Contributions.}
\begin{itemize}[leftmargin=1.2em,itemsep=0.2em]
  \item \textbf{Pretraining recipe} (\S\ref{sec:pretrain}): a two\mbox{-}stage pipeline of a shared scVI encoder over a $16{,}520$\mbox{-}gene cross\mbox{-}species ortholog vocabulary and a one\mbox{-}step Waddington\mbox{-}DiT dynamics backbone, trained on a 2.4M\mbox{-}cell mouse embryonic atlas spanning seven datasets and 88 timepoints, without any optimal\mbox{-}transport supervision.
  \item \textbf{Architecture} (\S\ref{sec:method}): a one\mbox{-}step action\mbox{-}conditioned residual with a Waddington\mbox{-}style decomposition into a potential gradient, an antisymmetric flow component, and a stochastic spread, with decoupled time embeddings for the gradient and rotational branches.
  \item \textbf{Time\mbox{-}transition transfer} (\S\ref{sec:downstream:weinreb}, \S\ref{sec:downstream:veres}, \S\ref{sec:downstream:fate}): used as fine\mbox{-}tuning initialization, the pretrained backbone improves per\mbox{-}target Sinkhorn $W_2$ on Weinreb $d4 / d6$ and on every Veres target $t1$ through $t7$ over matched scratch, PI\mbox{-}SDE, and PRESCIENT, and produces zero\mbox{-}shot clonal fate scores competitive with strong dynamic\mbox{-}OT baselines.
  \item \textbf{Perturbation embedding transfer} (\S\ref{sec:downstream:norman}): injected as the gene\mbox{-}state embedding inside GEARS, the pretrained dynamics representation reduces shared\mbox{-}vocabulary DE20 MSE on Norman Perturb\mbox{-}seq by $12.4\%$, evidence that pretrained developmental\mbox{-}trajectory dynamics encode differentiation primitives transferable to CRISPR\mbox{-}induced perturbation prediction.
\end{itemize}

\paragraph{Scope.}
This paper is about a controlled latent transition operator with a structured biological prior. We do not claim a general\mbox{-}purpose cell representation (\citealt{theodoris2023geneformer,cui2024scgpt} are stronger on that axis), we do not propose a knowledge\mbox{-}rich gene perturbation encoder (\citealt{alphacell2025,xcell2026} go further on that axis), and we do not benchmark against closed\mbox{-}source 4.9B\mbox{-}parameter systems whose pretraining data is proprietary. Our contribution is the structured one\mbox{-}step generator and its pretraining recipe; we expect the perturbation encoder to evolve in follow\mbox{-}up work without changing the residual backbone.

\section{Related work}
\label{sec:related_work}

\paragraph{Static perturbation predictors.}
A first family of methods treats perturbation response as a static map from a control population to a treated population. scGen adds a learned latent shift \citep{lotfollahi2019scgen}; CPA disentangles cellular state from a compositional perturbation embedding \citep{lotfollahi2023cpa}; GEARS couples a gene\mbox{-}graph prior with a graph neural network to predict multi\mbox{-}gene knockouts \citep{roohani2024gears}; and CellOT learns a neural optimal\mbox{-}transport map between control and perturbed marginals \citep{bunne2023cellot}. These methods model actions well, but discard temporal information: they fit one response per perturbation and do not answer queries at arbitrary time horizons. Chreode instead models the controlled transition $p(z_{t+\Delta}\mid z_t,\mathrm{do}(a))$ so that a trajectory at any $\Delta$ and any action is a single forward pass from the same backbone.

\paragraph{Dynamics from snapshot populations.}
Because scRNA\mbox{-}seq is destructive, cell dynamics are often inferred from unpaired snapshots. Waddington\mbox{-}OT estimates stochastic couplings between time points \citep{schiebinger2019wot}; TrajectoryNet links continuous normalizing flows with dynamic optimal transport \citep{tong2020trajectorynet}; PRESCIENT learns a stochastic potential landscape that can be integrated forward under interventions \citep{yeo2021prescient}; and dynamo reconstructs transcriptomic vector fields from velocity information \citep{qiu2022dynamo}. Recent flow\mbox{-}matching and Schr\"odinger\mbox{-}bridge variants further improve generative population transport: CFM with minibatch OT \citep{lipman2023flow,tong2023cfm}, BranchSBM extends this to branched multi\mbox{-}modal targets \citep{tang2025branchsbm}, CellFlow scales flow matching to phenotype modeling \citep{cellflow2025}, and CellStream learns dynamical OT\mbox{-}informed embeddings \citep{sha2025cellstream}. These methods use time well, but each forward query still requires iterative ODE/SDE or bridge integration, and each dataset is typically fit independently. We take PRESCIENT, BranchSBM, and CellFlow as our direct baselines; Chreode keeps their population\mbox{-}level view but amortizes the transition into one residual forward pass indexed by $\Delta$ and $a$, and pretrains a single backbone across all training trajectories.

\paragraph{Cell foundation models and concurrent world models.}
Large cell atlases can support transferable representations for cell annotation, integration, and gene\mbox{-}network reasoning \citep{theodoris2023geneformer,cui2024scgpt}, but neither Geneformer nor scGPT natively models a transition operator. Two concurrent cellular world models move toward the AI Virtual Cell agenda \citep{bunne2024aivirtualcell}. AlphaCell combines a knowledge\mbox{-}rich decoder with OT\mbox{-}CFM and targets perturbation response in unseen contexts \citep{alphacell2025}. X\mbox{-}Cell scales to 4.9B parameters and fits a diffusion language model on 25.6M perturbed transcriptomes \citep{xcell2026}. Both retain multi\mbox{-}step integration (OT\mbox{-}CFM or iterative diffusion denoising) and neither imposes an explicit Waddington\mbox{-}style biological prior on the transition operator. Because AlphaCell and X\mbox{-}Cell are closed\mbox{-}source and trained on proprietary perturbation compendia, we do not attempt a head\mbox{-}to\mbox{-}head performance comparison; instead we position Chreode through its structural differences, namely one\mbox{-}step inference and the Waddington residual decomposition, and we benchmark against open task\mbox{-}specific systems (BranchSBM, PRESCIENT, CellFlow, CellOT, scGen).

\paragraph{Waddington landscapes, flux, and one\mbox{-}step generators.}
Waddington's landscape remains the core abstraction for differentiation: cells move through basins of fate rather than arbitrary Euclidean space \citep{waddington1957}. Quantitative landscape theory adds that biological paths are shaped by both potential gradients and non\mbox{-}equilibrium flux, not by steepest descent alone \citep{wang2008landscape}. In parallel, world models in reinforcement learning learn latent state\mbox{-}action transitions for prediction and planning \citep{ha2018worldmodels,hafner2020dreamer}. We use the same abstraction for cells: a cytokine or a gene knockout is an action on a state, not merely metadata. Drifting Models \citep{deng2026drifting} recently show that a population\mbox{-}matching objective can train a generator whose pushforward is evaluated in one call rather than through many integration steps. Chreode combines this one\mbox{-}step amortization with cell\mbox{-}specific state\mbox{-}action semantics and a Waddington\mbox{-}style residual prior that is, to our knowledge, the first such decomposition in a cell foundation model.

Among the open task\mbox{-}specific systems we benchmark against in \S\ref{sec:downstream}, Chreode is the only entry that combines a single\mbox{-}step inference path, a pretrained backbone shared across downstream datasets, time and action conditioning, and a structured residual prior; a full per\mbox{-}method comparison on inference and structural axes is in Appendix~\ref{app:method_comparison}.

\section{Method}
\label{sec:method}

\begin{figure}[t]
  \centering
  \includegraphics[width=0.95\linewidth]{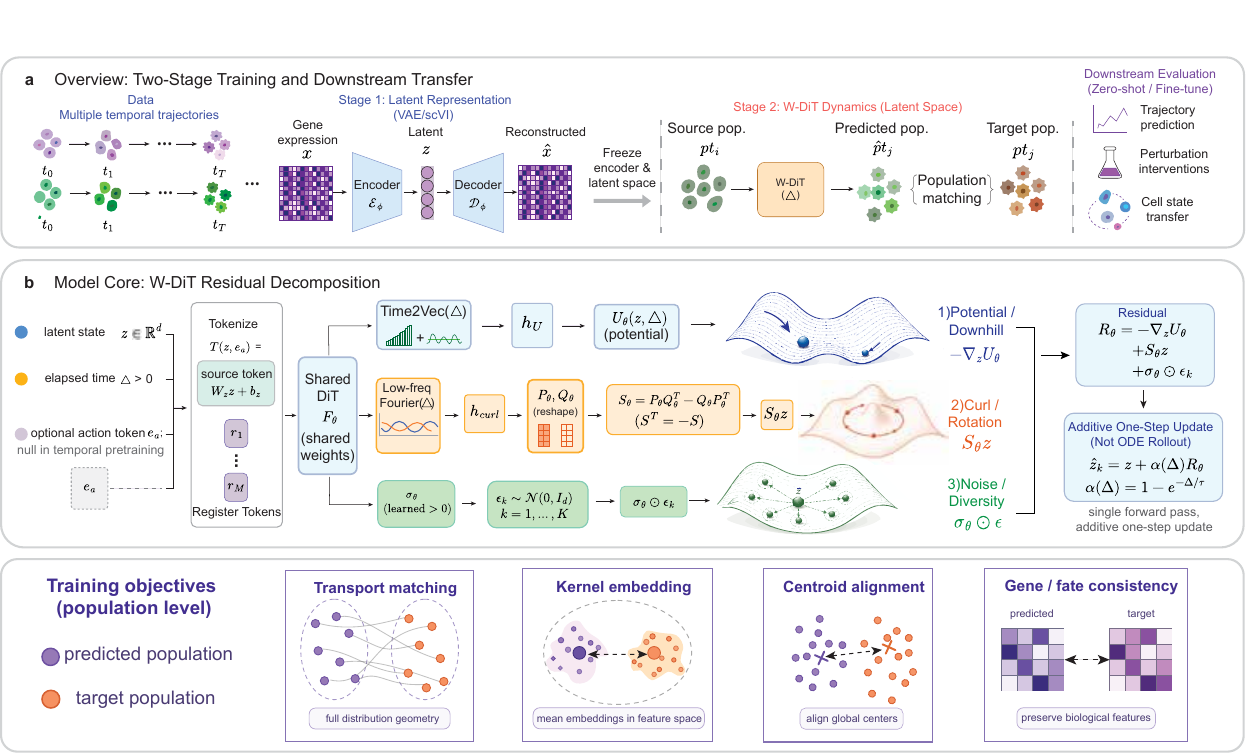}
  \caption{Chreode overview. \textbf{(a)} Two\mbox{-}stage pipeline: a Stage 1 scVI encoder $\mathcal E_\phi$ produces a frozen latent $z$, on top of which the Stage 2 Waddington\mbox{-}DiT learns a one\mbox{-}step latent transition $p_{t_i}\!\to\!\hat p_{t_j}$ matched against the observed $p_{t_j}$; the frozen backbone supports zero\mbox{-}shot and fine\mbox{-}tuned downstream transfer. \textbf{(b)} W\mbox{-}DiT residual decomposition: the state $z$, elapsed time $\Delta$, and optional action token $e_a$ are processed by a shared DiT, whose features feed three heads producing the potential gradient $-\nabla_z U_\theta$ (Time2Vec time code), the antisymmetric flow $S_\theta z$ with $S_\theta=P_\theta Q_\theta^\top-Q_\theta P_\theta^\top$ (bounded low\mbox{-}frequency Fourier time code), and the diagonal stochastic spread $\sigma_\theta\odot\epsilon_k$; the residual $R_\theta$ enters the one\mbox{-}step prediction $\hat z_k=z+\alpha(\Delta)R_\theta$, evaluated in a single forward pass per query. \textbf{(Bottom row)} Population\mbox{-}level training objectives jointly composed in Eq.~\eqref{eq:loss}.}
  \label{fig:overview}
\end{figure}

Chreode learns a controlled population transition $p(z_{t+\Delta}\mid z_t,\mathrm{do}(a))$ from snapshot data. Figure~\ref{fig:overview} summarizes the model. Panel (a) shows the two\mbox{-}stage training pipeline (Stage 1 latent encoder, Stage 2 W\mbox{-}DiT dynamics) and the downstream transfer modes; panel (b) shows the W\mbox{-}DiT residual decomposition into a potential\mbox{-}gradient term, an antisymmetric flow term, and a stochastic\mbox{-}spread term, all computed in a single forward pass; the bottom row shows the population\mbox{-}level training objectives that align predicted and target distributions. We first formalize the one\mbox{-}step prediction setup (\S\ref{sec:method:setup}), then motivate the three\mbox{-}component biological residual decomposition (\S\ref{sec:method:principles}), instantiate it as a DiT with decoupled time embeddings for the gradient and rotational branches (\S\ref{sec:method:architecture}), and train it with a population\mbox{-}matching loss that operates on unpaired snapshots (\S\ref{sec:method:objective}). The two design choices that distinguish Chreode from an unconstrained DiT residual head are the antisymmetric flow component and the decoupled time codes; we ablate both in \S\ref{sec:ablation}.

\subsection{Problem setup and notation}
\label{sec:method:setup}

Let $x\in\mathbb{R}^{G}$ denote a cell's gene expression vector over a fixed ortholog vocabulary of $G$ genes. A Stage~1 encoder $\mathcal{E}_\phi:x\mapsto z\in\mathbb{R}^{d}$ maps $x$ to a $d$\mbox{-}dimensional latent state with $d=128$ as the default ($d=64$ is used in the small\mbox{-}scale component validation of Appendix~\ref{app:component_validation}); a decoder $\mathcal{D}_\phi:z\mapsto x$ is available for gene\mbox{-}space evaluation. Stage~2 learns a one\mbox{-}step latent transition: given a source state $z\in\mathbb{R}^{d}$, an elapsed time $\Delta\ge 0$, an action $a$, and independent noise draws $\epsilon_{1},\ldots,\epsilon_{K}\in\mathbb{R}^{d}$, the model produces $K$ stochastic samples
\begin{equation}
  \hat z_{k}
  \;=\;
  z \;+\; \alpha(\Delta)\,R_\theta\!\left(z,\Delta,a,\epsilon_{k}\right),
  \qquad
  \alpha(\Delta)=1-\exp(-\Delta/\tau),
  \qquad
  \tau>0,
  \label{eq:one_step}
\end{equation}
that approximate the conditional population $p(z_{t+\Delta}\mid z_t=z,\mathrm{do}(a))$. The gate $\alpha(\Delta)$ enforces $\alpha(0)=0$ so that $\Delta=0$ collapses to the identity; $\tau$ is learnable and initialized from the median training $\Delta$. All of $U_\theta$, $S_\theta$, $\sigma_\theta$ depend on $(z,\Delta,a)$ and are the three components of the residual $R_\theta$.

\subsection{Design principles from Waddington's landscape}
\label{sec:method:principles}

Cell fate has been described for seventy years through Waddington's landscape metaphor \citep{waddington1957}: a cell rolls downhill on a developmental potential, while biological noise and rotational currents (cell\mbox{-}cycle progression, cyclic signaling) introduce directions that are not gradients of any scalar \citep{wang2008landscape}. This motivates parameterizing the residual as three interpretable biological effects rather than an unconstrained vector field:
\begin{equation}
  R_\theta(z,\Delta,a,\epsilon)
  \;=\;
  \underbrace{-\nabla_{z}U_{\theta}(z,\Delta,a)}_{\text{downhill potential}}
  \;+\;
  \underbrace{S_{\theta}(z,\Delta,a)\,z}_{\text{antisymmetric flow}}
  \;+\;
  \underbrace{\sigma_{\theta}(z,\Delta,a)\odot\epsilon}_{\text{stochastic spread}}.
  \label{eq:waddington_residual}
\end{equation}
Each term plays a distinct biological role: $-\nabla_{z}U_{\theta}$ drives cells toward low\mbox{-}potential basins of fate; the antisymmetry constraint $S_{\theta}^{\top}=-S_{\theta}$ makes $S_{\theta}z$ an in\mbox{-}tangent rotation rather than a gradient field, the simplest parameterization of the rotational current identified by quantitative landscape theory \citep{wang2008landscape}; and $\sigma_{\theta}\odot\epsilon$ captures intrinsic biological variability. We do not claim $S_{\theta}z$ is a Helmholtz curl in the strict differential\mbox{-}geometric sense; we use ``antisymmetric flow'' as the architectural realization of the gradient\mbox{-}plus\mbox{-}rotation dichotomy, and \S\ref{sec:ablation} shows it is the part of the residual that an unconstrained DiT head cannot easily express. Gated by $\alpha(\Delta)$ the residual is continuous in $\Delta$ by construction. We denote this parameterization the \emph{Waddington residual}.

\subsection{Architecture: a decoupled potential / antisymmetric DiT}
\label{sec:method:architecture}

We parameterize $U_{\theta}$, the low\mbox{-}rank factors of $S_{\theta}$, and $\sigma_{\theta}$ with a shared DiT \citep{peebles2023dit} feature extractor, but condition the potential and antisymmetric branches on \emph{different} time embeddings. The latent state token $W_{z}z + b_{z}$, an action token $e_{a}$ used by action\mbox{-}conditioned variants of the model, and $M$ learned register tokens are packed into a short sequence and processed by the shared DiT stack $F_{\theta}(\cdot, E)$, where $E$ is a time embedding injected through adaLN\mbox{-}zero conditioning, yielding source\mbox{-}slot features $h_{U}(z,a,\Delta)$ and $h_{\mathrm{curl}}(z,a,\Delta)$ for the two branches.

\paragraph{Why decouple the time embeddings.}
Long\mbox{-}$\Delta$ extrapolation is unstable when both branches share an unbounded Fourier code: $S_{\theta}z$ is norm\mbox{-}sensitive (unlike a gradient that integrates a scalar field), so high\mbox{-}frequency phases push the rotation through unseen angles at out\mbox{-}of\mbox{-}range $\Delta$. We therefore give the potential branch a flexible learnable Time2Vec code \citep{kazemi2019time2vec}, while restricting the antisymmetric branch to a bounded low\mbox{-}frequency periodic bank (periods $4,8,16,32,64,128$) so its rotation rates always interpolate training\mbox{-}seen frequencies. Functional forms are in Appendix~\ref{app:hyperparams}; the effect of decoupling is isolated in \S\ref{sec:ablation}.

\paragraph{Potential, antisymmetric, and noise heads.}
The potential is a scalar head $U_{\theta} = w_{U}^{\top} h_{U} + b_{U}$, and the downhill drift $-\nabla_{z} U_{\theta}$ is computed through autodiff. The antisymmetric head reshapes a linear projection of $h_{\mathrm{curl}}$ into two factors $P_{\theta}, Q_{\theta} \in \mathbb{R}^{d \times r}$ and forms
\begin{equation}
  S_{\theta}(z,a,\Delta)
  \;=\;
  P_{\theta}Q_{\theta}^{\top}\;-\;Q_{\theta}P_{\theta}^{\top},
  \qquad
  S_{\theta}^{\top}=-S_{\theta},
  \label{eq:antisym}
\end{equation}
so antisymmetry is exact by construction for any $(P_{\theta},Q_{\theta})$ output and the rank of $S_{\theta}$ is at most $2r$. The noise scale is elementwise positive, $\sigma_{\theta}=\mathrm{softplus}(\sigma_{\mathrm{raw}})+10^{-4}$. Substituting back into Eq.~\eqref{eq:one_step} yields $\hat z_{k} = z + \alpha(\Delta)\bigl[-\nabla_{z} U_{\theta} + S_{\theta} z + \sigma_{\theta} \odot \epsilon_{k}\bigr]$, a one\mbox{-}step additive update evaluated in a single forward pass per query.

\subsection{Population-level training objective}
\label{sec:method:objective}

Training does not require cell\mbox{-}to\mbox{-}cell pairs, which is essential because scRNA\mbox{-}seq is destructive and the same cell is never observed at two time points. For a transition $(t_{i},t_{j})$ and action $a$, we sample source cells $z\sim p_{t_{i}}$ and target cells $z'\sim p_{t_{j}}$ \emph{independently}, generate $\{\hat z_{k}\}_{k=1}^{K}$, and match the generated population to the target population. The loss combines kernel MMD, entropic Sinkhorn $W_{2}$, a stop\mbox{-}gradient drifting\mbox{-}field term adapted from \citet{deng2026drifting}, and a downhill regularizer on $U_{\theta}$:
\begin{equation}
  \mathcal L
  \;=\;
  \lambda_{\mathrm{mmd}}\,\mathcal L_{\mathrm{mmd}}
  \;+\;\lambda_{W_{2}}\,\mathcal L_{W_{2}}
  \;+\;\lambda_{\mathrm{drift}}\,\mathcal L_{\mathrm{drift}}
  \;+\;\lambda_{\mathrm{down}}\,\mathcal L_{\mathrm{down}}.
  \label{eq:loss}
\end{equation}
MMD and $W_{2}$ provide two complementary distributional matching signals; $\mathcal L_{\mathrm{drift}}$ enforces that the pushforward of $p_{t_{i}}$ under the residual matches $p_{t_{j}}$ at equilibrium (the anti\mbox{-}symmetry of the drift around $q=p$ is the invariant that makes this loss well behaved); and $\mathcal L_{\mathrm{down}}$ penalizes $U_{\theta}$ from moving uphill along the deterministic prediction.

For each trajectory $(t_{1},\ldots,t_{T})$ we sample training transitions uniformly from all ordered pairs $(t_{i},t_{j})$ with $i<j$, not only the largest $(t_{1},t_{T})$, which renders $\alpha(\Delta)$ identifiable (with a single $\Delta$ the $(\tau,\|R_\theta\|)$ pair is confounded). The action token $e_{a}$ is null throughout temporal pretraining; the perturbation transfer reported in \S\ref{sec:downstream:norman} uses the pretrained representation as a downstream gene\mbox{-}state embedding rather than via $e_{a}$. Full optimizer / schedule / multi\mbox{-}$\Delta$ details and rationale are in Appendix~\ref{app:hyperparams}.

\section{Pretraining}
\label{sec:pretrain}

Pretraining has two stages. Stage~1 fits a shared scVI \citep{lopez2018scvi} encoder--decoder over a fixed cross\mbox{-}species ortholog gene vocabulary; Stage~2 freezes the encoder and fits the Waddington\mbox{-}DiT dynamics backbone of \S\ref{sec:method} on population transitions across the training trajectories. Held\mbox{-}out downstream datasets are never seen in either stage.

\paragraph{Data.}
The pretraining corpus is a 2{,}477{,}217\mbox{-}cell mouse embryonic atlas aggregated from seven publicly available datasets and ten leaf trajectories (tome\mbox{-}mouse, GSE140802, GSE115943, E\mbox{-}MTAB\mbox{-}6967, GSE106340, GSE132188, GSE275562), spanning developmental times from 0 to 19 days post\mbox{-}fertilization across 88 sampled timepoints; the per\mbox{-}dataset cell counts and time ranges are in Appendix~\ref{app:pretrain_details}. Gene counts are harmonized to a single $\sim$33k\mbox{-}gene vocabulary and then restricted to 16{,}520 mouse--human one\mbox{-}to\mbox{-}one orthologs to enable cross\mbox{-}species transfer to human downstream datasets under the same input dimension. Expression is normalized with $\mathrm{normalize\_total}(10^{4})$ followed by $\log1p$.

\paragraph{Stage 1: shared scVI encoder.}
We train a single scVI model with latent dimension $d=128$ (a $d=64$ variant is reported in Appendix~\ref{app:component_validation}), hidden $512$, three layers, and a normal likelihood over $\log1p$\mbox{-}normalized expression. We treat \texttt{leaf\_dataset} as the technical batch covariate so that scVI removes inter\mbox{-}dataset and inter\mbox{-}lab effects, and we deliberately exclude developmental time from the batch covariate set: temporal variation is the biological signal that Stage~2 has to model and must therefore remain in the latent. The resulting encoder $\mathcal{E}_\phi$ and decoder $\mathcal{D}_\phi$ are frozen before Stage~2.

\paragraph{Stage 2: Waddington-DiT dynamics backbone.}
With Stage~1 frozen, we cache the latent cell matrix and train the dynamics backbone of \S\ref{sec:method:architecture} on population transitions $p_{t_{i}}\to p_{t_{j}}$ sampled uniformly from all ordered timepoint pairs within each leaf trajectory. We use AdamW ($\beta=(0.9,0.95)$, weight decay $0.01$), warmup cosine scheduling ($5\%$ warmup), batch size $512$, $K=8$ stochastic samples per source, and the loss of \eqref{eq:loss}; the action token $e_{a}$ is null throughout pretraining. The default model is the Small DiT (384 hidden, depth 12, heads 6, 4 register tokens); we also report a Tiny ablation (256 / 6 / 4 / 4) in Appendix~\ref{app:pretrain_details}.

\section{Downstream evaluation}
\label{sec:downstream}

We evaluate Chreode in two transfer modes, each on a distinct family of downstream tasks. As a \emph{fine\mbox{-}tuning initialization} (\S\ref{sec:downstream:weinreb}, \S\ref{sec:downstream:veres}), the pretrained Stage 2 dynamics backbone is used to initialize a downstream W\mbox{-}DiT trained on the held\mbox{-}out target dataset. As a \emph{gene\mbox{-}state embedding} (\S\ref{sec:downstream:norman}), the pretrained representation is injected into the existing GEARS perturbation predictor without modifying the GEARS training procedure. \S\ref{sec:downstream:fate} reports an additional zero\mbox{-}shot fate\mbox{-}prediction evaluation that uses the frozen pretrained backbone with no downstream training.

\paragraph{Protocol.}
For time\mbox{-}transition tasks, source cells are encoded into the frozen scVI\mbox{-}128 latent, predictions are generated with the dynamics backbone at the required $\Delta$, and metrics are computed in the shared latent after train\mbox{-}only standardization $\tilde z = (z - \mu_{\mathrm{train}}) / (\sigma_{\mathrm{train}} + 10^{-6})$ applied identically to source, target, and predicted populations. All baselines are trained on the same shared scVI\mbox{-}128 representation with the same train/test split, so comparisons are independent of representation choice. Mean and standard deviation are reported over 3 random seeds. Per\mbox{-}target Veres curves and decoded gene\mbox{-}space metrics are in Appendix~\ref{app:downstream_details}.

\subsection{Weinreb hematopoiesis}
\label{sec:downstream:weinreb}

\paragraph{Task.}
Predict held\mbox{-}out target populations at days 4 ($d4$) and 6 ($d6$) from a day\mbox{-}2 source on the Weinreb in\mbox{-}vitro lineage\mbox{-}tracing dataset \citep{yeo2021prescient}. The metric is Sinkhorn $W_{2}$ in the shared scVI\mbox{-}128 latent. Baselines are PI\mbox{-}SDE \citep{pisde2024}, PRESCIENT \citep{yeo2021prescient}, a matched\mbox{-}architecture scratch W\mbox{-}DiT (same backbone trained from random initialization on Weinreb only), and two sanity baselines (identity / source replay; linear time\mbox{-}delta).

\begin{table}[h]
  \centering
  \small
  \caption{Weinreb hematopoiesis. Lower is better. All rows are evaluated in the same shared scVI\mbox{-}128 latent with identical train/test splits and Sinkhorn $W_{2}$ implementation; mean $\pm$ std over 3 seeds where applicable.}
  \label{tab:results_weinreb}
  \begin{tabular}{lcc}
    \toprule
    Method & $W_{2}$ at $d4$ $\downarrow$ & $W_{2}$ at $d6$ $\downarrow$ \\
    \midrule
    \textbf{Chreode} (fine\mbox{-}tuned)   & \textbf{1.5133 $\pm$ 0.0757} & \textbf{1.6884 $\pm$ 0.0362} \\
    Scratch W\mbox{-}DiT (matched)            & 1.6387 $\pm$ 0.0106 & 2.0478 $\pm$ 0.0719 \\
    PI\mbox{-}SDE                             & 1.7452 $\pm$ 0.0219 & 1.8401 $\pm$ 0.0466 \\
    PRESCIENT                                  & 1.9096 $\pm$ 0.0124 & 1.8846 $\pm$ 0.0282 \\
    Identity / source replay                   & 2.6949              & 4.4553              \\
    Linear time\mbox{-}delta                   & 2.5326              & 3.7934              \\
    \bottomrule
  \end{tabular}
\end{table}

The pretrained backbone, used as fine\mbox{-}tuning initialization, achieves the lowest $W_{2}$ at both $d4$ and $d6$. The improvement over matched scratch holds across all three random seeds for both targets, isolating the value of pretraining over architecture and training budget alone. PI\mbox{-}SDE and PRESCIENT are trained on the same shared scVI\mbox{-}128 representation with the same split, so the head\mbox{-}to\mbox{-}head comparison is independent of representation choice.

\subsection{Veres islet differentiation}
\label{sec:downstream:veres}

\paragraph{Task.}
Predict each of seven target timepoints ($t = 1, \ldots, 7$) from a $t = 0$ source on the Veres islet differentiation dataset \citep{tang2025branchsbm}. We report the average per\mbox{-}target Sinkhorn $W_{2}$ over $t1$ through $t7$ and the $t7$ endpoint metric in the main table; the full per\mbox{-}target curve is in Appendix~\ref{app:downstream_details}.

\begin{table}[h]
  \centering
  \small
  \caption{Veres islet differentiation. Lower is better. The shared scVI\mbox{-}128 latent and the same train/test split are used for all rows.}
  \label{tab:results_veres}
  \begin{tabular}{lcc}
    \toprule
    Method & avg $W_{2}$ over $t = 1, \ldots, 7$ $\downarrow$ & $W_{2}$ at $t = 7$ $\downarrow$ \\
    \midrule
    \textbf{Chreode} (fine\mbox{-}tuned) & \textbf{2.6171} & \textbf{2.9132 $\pm$ 0.1704} \\
    Scratch W\mbox{-}DiT (matched)          & 2.8033          & 3.0892 $\pm$ 0.1144 \\
    PI\mbox{-}SDE                           & 2.8296          & 2.9490 $\pm$ 0.0980 \\
    PRESCIENT                                & 3.3219          & 3.8556 $\pm$ 0.4083 \\
    Identity / source replay                 & 6.9355          & 8.2324              \\
    Linear time\mbox{-}delta                 & 5.6000          & 6.2990              \\
    \bottomrule
  \end{tabular}
\end{table}

The pretrained backbone is the best row on every Veres target timepoint, including the $t7$ endpoint. The advantage holds at every $t \in \{1, \ldots, 7\}$ relative to PI\mbox{-}SDE, PRESCIENT, and matched scratch (full curve in Appendix~\ref{app:downstream_details}).

\subsection{Weinreb clonal fate prediction}
\label{sec:downstream:fate}

\paragraph{Task.}
Predict clonal lineage outcome on the Weinreb dataset following the PRESCIENT clonal evaluation protocol \citep{yeo2021prescient}: simulate stochastic endpoint trajectories from each $d2$ source cell, classify each predicted endpoint as Neutrophil/Monocyte/Other via a 20\mbox{-}NN classifier on $d6$ atlas cells, and score per\mbox{-}source fate as the predicted $\mathrm{Neu}/(\mathrm{Neu}+\mathrm{Mono})$ ratio. The metric is Pearson correlation $r_{\mathrm{masked}}$ between predicted and ground\mbox{-}truth fate ratios across clonally heldout cells. We use the frozen pretrained backbone in zero\mbox{-}shot mode (no Weinreb fine\mbox{-}tuning).

\begin{table}[h]
  \centering
  \small
  \caption{Weinreb clonal fate prediction. Higher is better. $r_{\mathrm{masked}}$ is Pearson correlation on cells with at least one classified endpoint prediction; $n_{\mathrm{with\_pred}}$ is the support.}
  \label{tab:results_fate}
  \begin{tabular}{lcc}
    \toprule
    Method & $r_{\mathrm{masked}}$ $\uparrow$ & $n_{\mathrm{with\_pred}}$ \\
    \midrule
    \textbf{Chreode} (zero\mbox{-}shot)            & \textbf{0.4680} & 183 \\
    scDiffEq \citep{scdiffeq2024}                  & 0.4627 & 296 \\
    moscot \citep{klein2025moscot}                 & 0.4624 & 265 \\
    WOT \citep{schiebinger2019wot}                 & 0.4593 & 265 \\
    Static\mbox{-}DiT control (zero\mbox{-}shot)   & 0.3825 & 335 \\
    \bottomrule
  \end{tabular}
\end{table}

The pretrained dynamics backbone in zero\mbox{-}shot mode produces fate scores competitive with strong dynamic\mbox{-}OT baselines (moscot, WOT) and the neural\mbox{-}SDE baseline scDiffEq. Replacing the dynamics\mbox{-}pretrained Stage 2 with a static\mbox{-}DiT control trained on reconstruction only drops the fate score by $0.085$, isolating that the temporal dynamics objective during pretraining is what produces this transfer.

\subsection{Norman Perturb-seq via GEARS embedding replacement}
\label{sec:downstream:norman}

\paragraph{Task and setup.}
Predict treated cell populations under CRISPRi single\mbox{-}gene knockouts from the same\mbox{-}cell control population on Norman Perturb\mbox{-}seq \citep{norman2019perturbseq} using the official GEARS predictor \citep{roohani2024gears}; our hypothesis is that developmental\mbox{-}atlas dynamics primitives (cells traversing differentiation manifolds) also describe CRISPR\mbox{-}induced state shifts. We test this by replacing the gene\mbox{-}state hidden embedding inside an otherwise\mbox{-}unchanged GEARS predictor with the corresponding Chreode representation. Four arms are compared under identical training (20 epochs, Norman split): the official GEARS embedding; raw scVI Stage 1 (VAE replace); the static\mbox{-}DiT (reconstruction only); and the dynamics\mbox{-}pretrained Stage 2 (Dynamics\mbox{-}DiT replace). We report the shared\mbox{-}vocabulary DE20 metric (top\mbox{-}20 differentially expressed genes selected within the gene set shared by our ortholog vocabulary and Norman).

\begin{table}[h]
  \centering
  \small
  \caption{Norman Perturb\mbox{-}seq. GEARS gene\mbox{-}state embedding replaced with three Chreode arms and an scVI baseline; identical GEARS training. Lower DE20 MSE / higher $r,\Delta r$ are better.}
  \label{tab:results_norman}
  \begin{tabular}{lccc}
    \toprule
    Run & DE20 MSE $\downarrow$ & DE20 $r$ $\uparrow$ & DE20 $\Delta r$ $\uparrow$ \\
    \midrule
    GEARS official                                       & 0.21208 & 0.79911 & 0.75931 \\
    GEARS + scVI replace                                 & 0.21262 & 0.79231 & 0.75556 \\
    GEARS + Static\mbox{-}DiT replace                    & 0.19358 & 0.80789 & 0.77007 \\
    GEARS + \textbf{Dynamics\mbox{-}DiT} replace         & \textbf{0.18580} & \textbf{0.81288} & \textbf{0.77135} \\
    \bottomrule
  \end{tabular}
\end{table}

The dynamics\mbox{-}pretrained representation reduces DE20 MSE by $12.4\%$ relative to unmodified GEARS. The progression scVI (neutral) $\to$ Static\mbox{-}DiT ($-8.7\%$) $\to$ Dynamics\mbox{-}DiT ($-12.4\%$) isolates the temporal\mbox{-}pretraining contribution: the Static\mbox{-}DiT $\to$ Dynamics\mbox{-}DiT step comes specifically from the dynamics objective encoding how states evolve along developmental trajectories. A velocity\mbox{-}consistency evaluation in CellStream\mbox{-}defined latents (Appendix~\ref{app:cellstream_velocity}) and a per\mbox{-}query timing comparison (Appendix~\ref{app:timing}) provide complementary evidence.

\section{Ablations}
\label{sec:ablation}

We ablate the architectural and training choices of \S\ref{sec:method} by re\mbox{-}pretraining Stage 2 with one component swapped per row, then fine\mbox{-}tuning each pretrained checkpoint on Weinreb and Veres under the same downstream protocol as Tab.~\ref{tab:results_weinreb}/Tab.~\ref{tab:results_veres} (3 seeds, 5000 epochs, shared scVI\mbox{-}128 latent). \emph{Group 1 (architecture)} tests whether the Waddington residual itself is the source of gain: \textbf{(a)} replacing \eqref{eq:waddington_residual} with an unconstrained DiT head $R_\theta=\mathrm{DiT}_\theta(z,\Delta,a,\epsilon)\in\mathbb{R}^{d}$ (no potential / antisymmetric / noise factorization); \textbf{(b)} tying both branches to the same Time2Vec embedding; \textbf{(c)} tying both to the bounded low\mbox{-}frequency Fourier code. \emph{Group 2 (training recipe)} holds the architecture fixed and tests whether the population objective and multi\mbox{-}$\Delta$ schedule are load\mbox{-}bearing: \textbf{(d)} single\mbox{-}$\Delta$ training (endpoint $(t_{1},t_{T})$ only); \textbf{(e)} dropping $\mathcal L_{\mathrm{drift}}$; \textbf{(f)} dropping $\mathcal L_{\mathrm{down}}$. Chreode is best on the benchmark\mbox{-}level averages (Weinreb avg and Veres avg, average rank $1.0$); single\mbox{-}$\Delta$ training is the most damaging swap, consistent with all\mbox{-}ordered multi\mbox{-}$\Delta$ training being necessary for identifying $\alpha(\Delta)$ at intermediate horizons. Full numbers (Table~\ref{tab:ablation}) and per\mbox{-}target details are in Appendix~\ref{app:ablation_details}; small\mbox{-}scale component validation is in Appendix~\ref{app:component_validation}.

\section{Scope and Future Work}
\label{sec:discussion}

Chreode pretrains a single backbone whose representation supports two distinct transfer modes (fine\mbox{-}tuning initialization for time\mbox{-}transition prediction; gene\mbox{-}state embedding for perturbation prediction inside GEARS). The Waddington residual decomposition makes one\mbox{-}forward\mbox{-}pass training feasible under a population\mbox{-}matching objective, and the developmental\mbox{-}atlas recipe makes the representation reusable across the time\mbox{-}transition / perturbation\mbox{-}prediction divide.

\paragraph{Why developmental dynamics transfers to perturbation prediction.}
The Norman improvement (\S\ref{sec:downstream:norman}) is, to our knowledge, the first empirical evidence that a dynamics representation pretrained purely on developmental atlases provides a trajectory prior for genetic perturbations on an unrelated dataset; we interpret this as both differentiation and CRISPR\mbox{-}induced shifts traversing cell\mbox{-}state manifolds along coherent trajectories, suggesting that scaling developmental\mbox{-}trajectory pretraining is a productive direction for foundation\mbox{-}scale perturbation prediction.

\paragraph{Scope and future work.}
Cross\mbox{-}species transfer relies on the $16{,}520$\mbox{-}gene ortholog vocabulary; adult human tissues are not evaluated. Time\mbox{-}transition transfer uses fine\mbox{-}tuning rather than strict zero\mbox{-}shot, the perturbation arm reuses GEARS, and pretraining scale is one to two orders below large world models on proprietary compendia. Future work (Appendix~\ref{app:limitations}): scaling the corpus, a fate\mbox{-}aware downstream criterion, and a gene\mbox{-}aware perturbation encoder end\mbox{-}to\mbox{-}end with the backbone.

\begin{ack}
Use unnumbered first level headings for the acknowledgments. All acknowledgments
go at the end of the paper before the list of references. Moreover, you are required to declare
funding (financial activities supporting the submitted work) and competing interests (related financial activities outside the submitted work).
More information about this disclosure can be found at: \url{https://neurips.cc/Conferences/2026/PaperInformation/FundingDisclosure}.

Do {\bf not} include this section in the anonymized submission, only in the final paper. You can use the \texttt{ack} environment provided in the style file to automatically hide this section in the anonymized submission.
\end{ack}



{
\small
\bibliographystyle{plainnat}
\bibliography{references}
}


\appendix

\section{Hyperparameters}
\label{app:hyperparams}

\subsection{Stage 1: shared scVI encoder}

\begin{table}[h]
  \centering
  \small
  \begin{tabular}{ll}
    \toprule
    Setting & Value \\
    \midrule
    Gene vocabulary          & 16{,}520 mouse--human 1:1 orthologs (Appendix~\ref{app:licenses}) \\
    Expression transform     & \texttt{normalize\_total}($10^{4}$) then $\log1p$ \\
    Likelihood               & Normal \\
    Latent dimension $d$     & $128$ (default); $64$ used in small-scale component validation \\
    Hidden dimension         & $512$ \\
    Depth                    & $3$ encoder / $3$ decoder layers \\
    Batch covariate          & \texttt{leaf\_dataset} ($10$ leaves) \\
    Timepoint batch correction & disabled (temporal signal retained for Stage 2) \\
    Optimizer                & Adam (scvi-tools default) \\
    Batch size               & $4096$ \\
    Epochs                   & $2$ ($\approx 4.95$M cell visits, validation plateaued) \\
    \bottomrule
  \end{tabular}
\end{table}

\subsection{Stage 2: Waddington-DiT dynamics backbone}

\begin{table}[h]
  \centering
  \small
  \begin{tabular}{ll}
    \toprule
    Setting & Value \\
    \midrule
    Backbone size            & DiT Small: hidden $384$, depth $12$, heads $6$, $4$ register tokens \\
    Curl low-rank dim $r$    & $16$ \\
    $U$ time embedding       & Time2Vec, $m=8$ periodic channels, bounded $\omega_{\max}=\pi$ \\
    Curl time embedding      & Bounded low-frequency Fourier, periods $(4,8,16,32,64,128)$ \\
    $\Delta$ normalization   & $\Delta_{\mathrm{scale}}=\tau_{\mathrm{init}}\cdot\ln 2$, per leaf trajectory \\
    Gate $\alpha(\Delta)$    & $1-\exp(-\Delta/\tau)$, $\tau$ learnable \\
    $\sigma_\theta$          & $\mathrm{softplus}(\sigma_{\mathrm{raw}})+10^{-4}$, elementwise \\
    Stochastic samples $K$   & $8$ (pretraining), $32$ (downstream evaluation) \\
    \midrule
    Loss                     & MMD + Sinkhorn $W_{2}$ + drifting field + downhill regularizer \\
    Loss weights             & $\lambda_{\mathrm{mmd}}{=}1$, $\lambda_{W_{2}}{=}1$, $\lambda_{\mathrm{drift}}{=}1$, $\lambda_{\mathrm{down}}{=}0.1$ \\
    Multi-$\Delta$ sampling  & uniform over all ordered $(t_{i},t_{j})$ pairs per leaf \\
    Optimizer                & AdamW, $\beta=(0.9,0.95)$, weight decay $0.01$ \\
    Learning-rate schedule   & warmup cosine, $5\%$ warmup fraction \\
    Base learning rate       & $3\times 10^{-4}$ \\
    Gradient clip            & $1.0$ \\
    Batch size               & $512$ source cells per step \\
    Training steps           & $3{,}356$ \\
    EMA                      & disabled (found not beneficial in component validation) \\
    Seed                     & single seed; downstream evaluation uses 3 seeds of data/noise draws \\
    \bottomrule
  \end{tabular}
\end{table}

\paragraph{Multi-$\Delta$ training rationale.}
For each leaf trajectory with timepoints $(t_{1},\ldots,t_{T})$, training transitions are sampled uniformly from all ordered pairs $(t_{i},t_{j})$ with $i<j$, not only from the largest $(t_{1},t_{T})$. This is what renders $\alpha(\Delta)$ identifiable from data: with a single observed $\Delta$, the pair $(\tau,\|R_\theta\|)$ is confounded because increasing $\tau$ and rescaling $R_\theta$ produces the same one\mbox{-}step prediction; only with multiple $\Delta$'s does the shape of $\alpha(\Delta)$ become uniquely pinned. The action token $e_{a}$ is null throughout pretraining on mouse embryonic atlas trajectories, so the model learns action\mbox{-}free temporal dynamics; the perturbation transfer reported in \S\ref{sec:downstream:norman} uses the pretrained dynamics representation as a downstream gene\mbox{-}state embedding rather than via $e_{a}$. The full ablation against single\mbox{-}$\Delta$ training is in Appendix~\ref{app:ablation_details}.

\subsection{Compute resources}

Stage 1 scVI training uses $1$ GPU. Stage 2 dynamics pretraining uses $1$ GPU. Each downstream evaluation run uses a single GPU.

\subsection{Downstream evaluation}

For each downstream task, the frozen encoder maps source cells to the scVI latent, and the frozen dynamics backbone produces $K=32$ stochastic samples per source cell at the required $(\Delta,a)$. We report mean and standard deviation across 3 random seeds that control data splits and evaluation noise draws; the pretrained backbone itself is a single checkpoint.

\paragraph{Metric definitions.}
Sinkhorn $W_{2}$ uses entropic regularization $\varepsilon=0.1$ and $100$ Sinkhorn iterations. MMD uses a sum of RBF kernels over bandwidths $\{0.001, 0.01, 0.1, 1, 10, 100\}$. Fate Pearson $r_{\mathrm{masked}}$ on Weinreb uses the clonal lineage ground truth reconstructed from publicly released Klein-lab annotations, evaluated only on clones with at least two daughter cells. The shared-vocabulary DE20 metric on Norman is computed by selecting the top-20 differentially expressed genes per condition within the gene set shared by our pretraining ortholog vocabulary and the Norman vocabulary, then computing per-condition mean squared error and Pearson correlation between predicted and observed log fold changes on those 20 genes.

\section{Pretraining details}
\label{app:pretrain_details}

\paragraph{Pretraining corpus breakdown.}
Table~\ref{tab:pretrain_data} reports the per\mbox{-}dataset cell counts and time ranges of the 2{,}477{,}217\mbox{-}cell pretraining corpus referenced in \S\ref{sec:pretrain}. The seven datasets cover ten leaf trajectories spanning developmental times from 0 to 19 days post\mbox{-}fertilization across 88 sampled timepoints.

\begin{table}[h]
  \centering
  \caption{Pretraining corpus. 2.48M cells, seven datasets, ten leaf trajectories, 16.5k ortholog genes.}
  \label{tab:pretrain_data}
  \small
  \begin{tabular}{lrrl}
    \toprule
    Dataset family & h5ad files & Cells & Time range (days) \\
    \midrule
    tome\mbox{-}mouse & 19 & 1{,}658{,}968 & 3.5 -- 13.5 \\
    GSE140802 (3 leaves: inVitro, inVivo, cytokine) & 9 & 378{,}135 & 2.0 -- 16.0 \\
    GSE115943 (C1 + C2) & 80 & 181{,}188 & 0.0 -- 19.0 \\
    E\mbox{-}MTAB\mbox{-}6967 & 9 & 130{,}006 & 6.5 -- 8.5 \\
    GSE106340 & 11 & 68{,}339 & 0.0 -- 17.0 \\
    GSE132188 & 4 & 37{,}977 & 12.5 -- 15.5 \\
    GSE275562 & 3 & 22{,}604 & 14.5 -- 16.5 \\
    \midrule
    \textbf{Total} & \textbf{135} & \textbf{2{,}477{,}217} & \\
    \bottomrule
  \end{tabular}
\end{table}

\paragraph{Ortholog vocabulary.}
The cross-species ortholog table is built from Ensembl BioMart, filtered to mouse--human one-to-one orthologs with \texttt{ortholog\_confidence}$=1$. This yields $16{,}520$ gene pairs. The mouse vocabulary of the pretraining atlas ($\sim$33k genes) is intersected with this table, yielding $16{,}485$ genes present in both; the remaining 35 genes are mapped via symbol aliases.

\paragraph{Stage 1 reconstruction quality.}
Stage 1 is trained for $2$ epochs ($1{,}678$ optimizer steps, $\approx 4.95$M cell visits) until the validation objective plateaued. The selected checkpoint reaches a held-out reconstruction Pearson of $0.586$, validation negative ELBO of $0.774$, latent mean $-0.017$, and latent standard deviation $0.84$.

\paragraph{Stage 2 training curves.}
After $3{,}356$ training steps, the last-10-batch median of the loss components is MMD $0.037$, Sinkhorn $W_{2}$ $31.5$, drift $8.68$, and downhill $\approx\!0$, decreasing monotonically from first-10 medians of $0.21 / 81.2 / 8.83 / \approx\!0$ respectively. The held-out temporal evaluation reaches Sinkhorn $W_{2}=37.5$, MMD $=0.052$, and $\Delta$ Pearson $=0.551$. No NaN or divergence was observed.

\section{Downstream evaluation details}
\label{app:downstream_details}

\paragraph{Weinreb.}
Per-timepoint $W_{2}$ at $d4$ and $d6$ for the Chreode (fine-tuned), scratch, PI-SDE, PRESCIENT, and sanity-baseline comparison, plus fate Pearson breakdowns and per-clonal-size fate predictions for the zero-shot fate evaluation.

\paragraph{Veres.}
Per-timepoint $W_{2}$ across $t=1,\ldots,7$ for the Chreode (fine-tuned), scratch, PI-SDE, and PRESCIENT comparison.

\paragraph{Norman.}
Per-condition DE20 mean squared error and Pearson correlation breakdowns for the four GEARS arms (official, VAE replace, Static-DiT replace, Dynamics-DiT replace), and the subset of conditions whose orthologous gene coverage exceeds $80\%$.

\section{Small-scale component validation}
\label{app:component_validation}

Before pretraining at scale, we validated the Waddington-DiT residual architecture and the training recipe on small-scale from-scratch benchmarks on Weinreb ($d2\to d6$) and Veres (7 timepoints). These numbers do not count as foundation-model contributions and are reported here as engineering evidence that each architectural decision (decoupled time embeddings, additive vs Cayley update, multi-$\Delta$ training, loss balancer) produces the expected effect at small scale.

\paragraph{Selected architectural decisions validated at small scale.}
(i) Multi-$\Delta$ training improves intermediate-timepoint Weinreb $W_{2}$ over single-$\Delta$ training by a large margin. (ii) Bounded low-frequency Fourier time features in the antisymmetric branch produce stable long-$\Delta$ behavior, whereas tying both branches to a shared unbounded Fourier code is unstable beyond the training horizon. (iii) Using Time2Vec in the potential branch while keeping bounded low-frequency Fourier in the antisymmetric branch gives the best tradeoff between in-distribution accuracy and long-$\Delta$ stability. Full ablation tables are available in the anonymized repository.

\section{Licenses for existing assets}
\label{app:licenses}

\begin{table}[h]
  \centering
  \small
  \caption{Pretraining and downstream datasets used in this paper. All datasets are publicly available.}
  \begin{tabular}{@{}>{\raggedright\arraybackslash}p{0.20\linewidth}>{\raggedright\arraybackslash}p{0.30\linewidth}>{\raggedright\arraybackslash}p{0.42\linewidth}@{}}
    \toprule
    Dataset & Accession / source & Notes \\
    \midrule
    E-MTAB-6967            & ArrayExpress E-MTAB-6967 & mouse gastrulation \\
    GSE106340              & GEO GSE106340 & mouse embryo \\
    GSE115943 (C1, C2)     & GEO GSE115943 & mouse embryo, two replicates \\
    GSE132188              & GEO GSE132188 & mouse embryo \\
    GSE140802              & GEO GSE140802 & inVitro / inVivo / cytokine leaves \\
    GSE275562              & GEO GSE275562 & mouse embryo \\
    tome-mouse             & \citet{qiu2022dynamo} data release & mouse development atlas \\
    \midrule
    Weinreb hematopoiesis  & GEO GSE140802 subset & as released with \citet{yeo2021prescient}; used for both time-transition and fate evaluation \\
    Veres islet            & as released with \citet{tang2025branchsbm} & 7-timepoint differentiation \\
    Norman Perturb-seq     & GEO GSE133344 & \citep{norman2019perturbseq} \\
    CellStream EMT, MOSTA  & shipped with \citet{sha2025cellstream} & appendix velocity-consistency evaluation \\
    \bottomrule
  \end{tabular}
\end{table}

Baseline software: PRESCIENT \citep{yeo2021prescient} under its GitHub license; BranchSBM \citep{tang2025branchsbm} under its GitHub license; CellFlow \citep{cellflow2025} under its release license; CellOT \citep{bunne2023cellot} and scGen \citep{lotfollahi2019scgen} under their respective open-source licenses. The scVI framework \citep{lopez2018scvi} is used under the scvi-tools BSD-3 license.

\section{Broader impacts}
\label{app:broader_impacts}

\paragraph{Positive impacts.}
A cell world model that predicts controlled state transitions in one forward pass can reduce the cost and latency of in-silico screens for drug and genetic perturbations, allowing wet-lab experiments to be more targeted and accelerating the prioritization of therapeutic candidates. By reusing a single pretrained backbone across developmental and perturbational downstreams, Chreode also consolidates a landscape of previously task-specific systems, which makes reproducibility, auditability, and iterative refinement easier for the community.

\paragraph{Potential negative impacts.}
Any model capable of predicting cellular responses to perturbations shares a generic dual-use concern: a sufficiently accurate predictor could in principle be used to nominate interventions with harmful biological effects. We note that Chreode is trained on developmental transcriptomic atlases rather than on pathogen, toxin, or human-clinical data, so the kinds of interventions it is exposed to at pretraining time are developmental signals rather than cytotoxic or infection-related perturbations. Readers interested in dual-use policy for perturbation-prediction models are referred to ongoing community discussions around responsible release of biological foundation models.

\paragraph{Release.}
We release the Chreode codebase and raw downstream predictions under an open research license. Pretrained weights are not released with this submission.

\section{Inference\mbox{-}cost comparison}
\label{app:timing}

Table~\ref{tab:timing} reports the per\mbox{-}query inference cost of Chreode and the dominant multi\mbox{-}step baselines benchmarked in \S\ref{sec:downstream}, measured on a single NVIDIA A100.

\begin{table}[h]
  \centering
  \small
  \caption{Per\mbox{-}query inference cost on a single NVIDIA A100, fp32, batch size $1$. ``NFE'' counts the number of network forward passes per (source cell, $\Delta$, action) query. Wall\mbox{-}clock and FLOPs are reported as median over $60$ Weinreb $d2{\to}d6$ queries with $8$ warmup queries discarded; FLOPs from \texttt{torch.profiler}.}
  \label{tab:timing}
  \begin{tabular}{lccc}
    \toprule
    Method            & NFE / query     & ms / query     & GFLOPs / query \\
    \midrule
    PRESCIENT         & ${\sim}88$      & $194.33$       & $0.07$ \\
    CellFlow          & 50--100 ODE     & $410.36$       & --- \\
    \midrule
    \textbf{Chreode}  & \textbf{1}      & \textbf{$64.79$}   & \textbf{$0.90$} \\
    \bottomrule
  \end{tabular}
\end{table}

Chreode resolves a query in a single DiT\mbox{-}Small forward pass and is therefore $3.0\times$ faster than PRESCIENT and $6.3\times$ faster than CellFlow at single\mbox{-}query latency, even though its forward pass includes the autograd backprop required by the $-\nabla_z U_\theta$ head. The result projects directly onto screen scale: a $10^{8}$\mbox{-}query screen on a single A100 takes approximately $1{,}800$ hours for Chreode versus $5{,}398$ hours for PRESCIENT and $11{,}399$ hours for CellFlow. CellFlow GFLOPs are not reported because its JAX/diffrax solve is outside the PyTorch profiler.

\section{Velocity consistency in CellStream\mbox{-}defined latent spaces}
\label{app:cellstream_velocity}

As an additional check on whether the pretrained backbone produces smooth, locally\mbox{-}consistent velocity fields, we evaluate Chreode in the latent spaces defined by the CellStream pretrained encoder \citep{sha2025cellstream} on three datasets shipped with that codebase. We use Chreode (fine\mbox{-}tuned) as the dynamics arm, with the pretrained Stage 2 backbone fine\mbox{-}tuned on each dataset's CellStream latent for the same number of epochs as the local CellStream comparison. The metric is kNN\mbox{-}20 velocity consistency (VC) and Sinkhorn $W_2$ at the endpoint timepoint, both computed in CellStream's own latent space.

\begin{table}[h]
  \centering
  \small
  \caption{Velocity consistency in CellStream\mbox{-}defined latent spaces. VC is kNN\mbox{-}20 cosine consistency of predicted local velocity ($\uparrow$); endpoint $W_{2}$ is Sinkhorn $W_{2}$ at the dataset's last observed timepoint ($\downarrow$). Mean over 3 seeds where applicable.}
  \label{tab:cellstream_appendix}
  \begin{tabular}{llcc}
    \toprule
    Dataset & Method                       & VC kNN20 $\uparrow$    & endpoint $W_{2}$ $\downarrow$ \\
    \midrule
    EMT     & CellStream (pretrained)     & 0.9743                 & --                            \\
    EMT     & \textbf{Chreode} (fine\mbox{-}tuned)    & \textbf{0.9983 $\pm$ 0.0002} & --                      \\
    \midrule
    MOSTA   & CellStream (pretrained)     & 0.9638                 & 0.0358 $\pm$ 0.0025          \\
    MOSTA   & \textbf{Chreode} (fine\mbox{-}tuned)    & \textbf{0.9992 $\pm$ 0.0003} & \textbf{0.0355 $\pm$ 0.0009} \\
    \bottomrule
  \end{tabular}
\end{table}

The pretrained backbone, used as fine\mbox{-}tuning initialization in the CellStream\mbox{-}defined latent, produces velocity fields whose local consistency exceeds the CellStream pretrained model on EMT and MOSTA, and matches CellStream's MOSTA endpoint $W_2$. We treat this as evidence that the dynamics representation captures meaningful local direction, complementary to the in\mbox{-}distribution scVI\mbox{-}128 results in \S\ref{sec:downstream}.

\section{Scope notes}
\label{app:limitations}

\paragraph{Pretraining corpus is mouse\mbox{-}embryonic.}
Cross\mbox{-}species transfer to human downstream datasets is mediated only by the $16{,}520$\mbox{-}gene mouse--human ortholog vocabulary; we do not pretrain on any human transcriptomes. A richer cross\mbox{-}species pretraining schedule that includes adult human tissues is left to future work.

\paragraph{Time\mbox{-}transition transfer uses fine\mbox{-}tuning, not zero\mbox{-}shot.}
Weinreb and Veres results in \S\ref{sec:downstream:weinreb}, \S\ref{sec:downstream:veres} use the pretrained backbone as initialization for downstream fine\mbox{-}tuning. The clonal fate evaluation in \S\ref{sec:downstream:fate} uses the frozen pretrained backbone in zero\mbox{-}shot mode. Whether a single training recipe can cover both regimes simultaneously, including a clone\mbox{-}aware downstream criterion, is an open question.

\paragraph{Perturbation transfer reuses GEARS rather than a standalone operator.}
The Norman result in \S\ref{sec:downstream:norman} comes from injecting the pretrained representation as the gene\mbox{-}state embedding inside an unmodified GEARS predictor. We do not present a standalone perturbation operator from the pretrained backbone in this paper; designing a gene\mbox{-}aware perturbation encoder that composes end\mbox{-}to\mbox{-}end with the dynamics backbone is a natural follow\mbox{-}up.

\paragraph{Scale is below concurrent large cellular world models.}
Our pretraining scale (2.4M cells, Small DiT, of order $10^{7}$ parameters) is one to two orders of magnitude below concurrent large cellular world models such as AlphaCell \citep{alphacell2025} and X\mbox{-}Cell \citep{xcell2026}, which train on proprietary perturbation compendia. Whether the present transfer modes scale to 100M\mbox{-}cell developmental atlases is an open empirical question.

\paragraph{No comparison to representation\mbox{-}only foundation models.}
We cite Geneformer and scGPT \citep{theodoris2023geneformer,cui2024scgpt} but do not benchmark against them in this paper, because they do not expose a transition operator and require a nontrivial adapter head to be evaluated on our downstream regimes. Comparison to representation\mbox{-}only foundation models is orthogonal to the present contribution and is left to follow\mbox{-}up work.

\paragraph{Latent\mbox{-}space evaluation by default.}
$W_{2}$, MMD, and energy\mbox{-}distance metrics are computed in the shared scVI\mbox{-}128 latent; gene\mbox{-}space metrics for Norman use the Stage 1 decoder. Additional gene\mbox{-}level evaluation panels are reported in \S\ref{app:downstream_details}.

\paragraph{No uncertainty quantification yet.}
The model exposes a per\mbox{-}dimension noise scale $\sigma_{\theta}(z,\Delta,a)$, but we do not currently calibrate or report posterior credible intervals on downstream predictions. Calibration of the stochastic spread against held\mbox{-}out replicate variance is a natural follow\mbox{-}up.

\section{Method comparison details}
\label{app:method_comparison}

Table~\ref{tab:method_axes} expands on the inference and structural differences between Chreode and the open task\mbox{-}specific systems we benchmark against in \S\ref{sec:downstream}. We report each method along five axes: per\mbox{-}query inference cost, whether a pretrained backbone is shared across downstream datasets, whether the operator is conditioned on elapsed time $\Delta$, whether it accepts an action input $a$, and what architectural inductive bias (if any) the residual / drift carries.

\begin{table}[h]
\centering
\small
\caption{Method comparison on inference and structural axes. ``Inference cost'' uses each method's typical released setting; ``architectural residual prior'' refers to architectural inductive bias rather than data preprocessing.}
\label{tab:method_axes}
\resizebox{\linewidth}{!}{
\begin{tabular}{lccccc}
\toprule
Method     & Inference            & Pretrained        & Time\mbox{-}cond. & Action\mbox{-}cond. & Architectural \\
           & cost                 & backbone          & ($\Delta$)        & ($a$)               & residual prior \\
\midrule
PRESCIENT  & ${\sim}88$ Euler steps & per dataset    & yes               & limited             & potential landscape \\
BranchSBM  & multi\mbox{-}step SDE  & per dataset    & yes               & no                  & none \\
CellFlow   & 40--100 ODE steps    & per dataset      & yes               & yes (control)       & none \\
scGen      & 1 step (no time)     & per dataset      & no                & yes                 & latent additive \\
CPA        & 1 step (no time)     & per dataset      & no                & yes                 & additive composition \\
\midrule
\textbf{Chreode} & \textbf{1 step} & \textbf{2.4M cells} & \textbf{$\alpha(\Delta)$ gate} & \textbf{token} & \textbf{antisymmetric flow + potential} \\
\bottomrule
\end{tabular}}
\end{table}

Chreode is the only entry that combines (i) a single\mbox{-}step inference path, (ii) a pretrained backbone shared across all downstream datasets rather than retrained per dataset, (iii) joint time and action conditioning, and (iv) a structured residual prior that decomposes the velocity field into a potential gradient, an antisymmetric flow, and a stochastic spread. Among the multi\mbox{-}step rows, PRESCIENT carries the closest structural inductive bias (a potential landscape) but no antisymmetric component and no shared pretraining; among the one\mbox{-}step rows, scGen and CPA add latent\mbox{-}additive priors but discard time entirely. We do not include the closed\mbox{-}source AlphaCell and X\mbox{-}Cell systems in this table because their architectural details and inference protocols are proprietary.

\section{Ablation details}
\label{app:ablation_details}

This appendix expands on the architectural and training ablations summarized in \S\ref{sec:ablation}. Each row of Table~\ref{tab:ablation} is a separately pretrained Stage 2 checkpoint with one component swapped relative to the selected Chreode configuration; Stage 1 (the scVI encoder) is held fixed across all ablation rows. Each pretrained checkpoint is then fine\mbox{-}tuned on Weinreb and Veres under the same downstream protocol as Tab.~\ref{tab:results_weinreb}/Tab.~\ref{tab:results_veres} (3 seeds, 5000 epochs, shared scVI\mbox{-}128 latent), so absolute Sinkhorn $W_{2}$ values are directly comparable to the main downstream tables. The headline columns are benchmark\mbox{-}level averages (Weinreb avg over $\{d4,d6\}$ and Veres avg over $t1$ through $t7$) plus the average rank across the two; per\mbox{-}target detail is provided in the last two columns.

\begin{table}[h]
  \centering
  \caption{Architectural and training ablations under matched downstream fine\mbox{-}tuning. Headline columns are benchmark\mbox{-}level averages (Weinreb avg over $d4/d6$, Veres avg over $t1$ through $t7$) plus the average rank across both. Lower is better.}
  \label{tab:ablation}
  \small
  \resizebox{\linewidth}{!}{
  \begin{tabular}{lccccc}
    \toprule
    Variant & Weinreb avg $\downarrow$ & Veres avg $\downarrow$ & Avg rank $\downarrow$ & Weinreb $d4$ $\downarrow$ & Weinreb $d6$ $\downarrow$ \\
    \midrule
    \textbf{Chreode (selected)}                 & \textbf{1.6008} & \textbf{2.6171} & \textbf{1.0} & \textbf{1.5133} & 1.6884 \\
    \midrule
    \multicolumn{6}{l}{\emph{G1: architecture}} \\
    Unconstrained DiT residual                  & 1.8922 & 3.4030 & 2.0 & 2.1329 & \textbf{1.6516} \\
    Tied time embeddings (both Time2Vec)        & 1.9073 & 3.7139 & 5.0 & 2.0868 & 1.7278 \\
    Tied time embeddings (both low\mbox{-}freq) & 1.9416 & 3.5298 & 4.0 & 2.1254 & 1.7578 \\
    \midrule
    \multicolumn{6}{l}{\emph{G2: training recipe}} \\
    Single\mbox{-}$\Delta$ training       & 2.0579 & 3.7938 & 7.0 & 2.1313 & 1.9845 \\
    Without $\mathcal L_{\mathrm{drift}}$ & 1.8974 & 3.6258 & 3.5 & 2.0878 & 1.7071 \\
    Without $\mathcal L_{\mathrm{down}}$  & 1.9835 & 3.6793 & 5.5 & 2.1015 & 1.8655 \\
    \bottomrule
  \end{tabular}}
\end{table}

\paragraph{Reading the headline.}
The selected Chreode is the best row on both benchmark\mbox{-}level aggregates (Weinreb avg $1.6008$, Veres avg $2.6171$) and obtains the best average rank across the two. The Veres advantage is the most decisive: every ablation row exceeds Chreode's Veres average by $30$--$45\%$ ($3.40$--$3.79$ vs $2.6171$), which is consistent with the design target of multi\mbox{-}target temporal transfer.

\paragraph{Reading G1 (architecture).}
\emph{Unconstrained DiT residual} (no potential / antisymmetric / noise factorization) is the closest competitor at the Weinreb $d6$ endpoint ($1.6516$ vs $1.6884$, within seed variance: std $0.21$ vs $0.04$), but its Weinreb $d4$ ($2.1329$ vs $1.5133$, $+41\%$) and Veres average ($3.4030$, $+30\%$) are substantially worse: the unstructured head can fit a single endpoint but cannot match the structured residual on early targets or multi\mbox{-}target Veres, isolating the value of the Waddington decomposition for multi\mbox{-}$\Delta$ transfer. \emph{Tied Time2Vec} (both branches share the unbounded Time2Vec code) increases Veres average to $3.7139$ ($+42\%$); we attribute this to the antisymmetric branch becoming unstable at long $\Delta$ when its rotation rates drift through unseen frequencies, consistent with the architectural rationale in \S\ref{sec:method:architecture}. \emph{Tied low\mbox{-}frequency Fourier} (both branches share the bounded code) reaches Veres average $3.5298$ ($+35\%$): a single bounded code is not flexible enough to match the long monotone developmental schedule that the potential branch needs.

\paragraph{Reading G2 (training recipe).}
\emph{Single\mbox{-}$\Delta$ training} is the most damaging swap in either group (Weinreb avg $2.0579$, Veres avg $3.7938$, average rank $7.0$). This matches the design rationale that all\mbox{-}ordered multi\mbox{-}$\Delta$ training is necessary for identifying the time gate $\alpha(\Delta)$ at intermediate horizons (Appendix~\ref{app:hyperparams}, ``Multi\mbox{-}$\Delta$ training rationale''): when only the largest $\Delta$ is observed at training time, the model has no signal to disambiguate the gate shape from the residual norm. \emph{Without $\mathcal L_{\mathrm{drift}}$} (Weinreb avg $1.8974$, Veres avg $3.6258$) and \emph{without $\mathcal L_{\mathrm{down}}$} (Weinreb avg $1.9835$, Veres avg $3.6793$) both underperform the selected Chreode by $20$--$40\%$ on the aggregates; removing $\mathcal L_{\mathrm{down}}$ hurts Weinreb more (Weinreb $d6$ $1.8655$ vs $1.6884$), while removing $\mathcal L_{\mathrm{drift}}$ hurts Veres more, indicating dataset\mbox{-}dependent value rather than a uniform monotone effect, but both regularizers help on average and we keep them in the selected configuration.

\paragraph{Detailed Weinreb metrics, selected Chreode vs unconstrained DiT.}
Because Weinreb $d6$ full\mbox{-}latent $W_{2}$ is close between the selected configuration and the unconstrained DiT (within seed variance), Table~\ref{tab:weinreb_detailed_ablation} reports additional Weinreb metrics from the same fine\mbox{-}tuned runs: Sinkhorn $W_{1}$ and $W_{2}$ on the top\mbox{-}$2$ principal components, full\mbox{-}latent $W_{1}$, BranchSBM\mbox{-}style fixed\mbox{-}bandwidth MMD, and our unbiased median\mbox{-}heuristic MMD. At $d4$ the selected configuration is better on every reported metric. At $d6$ the selected configuration is better on top\mbox{-}$2$ $W_{1}/W_{2}$ and both MMD statistics, while unconstrained DiT is slightly better on full\mbox{-}latent $W_{1}/W_{2}$ but with $5\times$ larger seed variance. The combined picture supports the headline reading: the structured residual gives more stable distribution matching at $d4$ and on low\mbox{-}dimensional / kernel metrics at $d6$, while a single full\mbox{-}latent $d6$ endpoint metric is not enough to favor either side.

\begin{table}[h]
  \centering
  \small
  \caption{Detailed Weinreb metrics for selected Chreode vs unconstrained DiT under fine\mbox{-}tuning, mean $\pm$ std over 3 seeds. Lower is better.}
  \label{tab:weinreb_detailed_ablation}
  \resizebox{\linewidth}{!}{
  \begin{tabular}{llccccc}
    \toprule
    Method & Day & Top\mbox{-}$2$ $W_{1}$ $\downarrow$ & Top\mbox{-}$2$ $W_{2}$ $\downarrow$ & MMD full $\downarrow$ & Full $W_{1}$ $\downarrow$ & Full $W_{2}$ $\downarrow$ \\
    \midrule
    \textbf{Chreode}    & $d4$ & \textbf{0.046$\pm$0.005} & \textbf{0.079$\pm$0.008} & \textbf{0.0074$\pm$0.0003} & \textbf{1.286$\pm$0.031} & \textbf{1.513$\pm$0.076} \\
    Unconstrained DiT   & $d4$ & 0.102$\pm$0.009 & 0.147$\pm$0.018 & 0.0349$\pm$0.0027 & 1.860$\pm$0.005 & 2.133$\pm$0.007 \\
    \midrule
    \textbf{Chreode}    & $d6$ & \textbf{0.043$\pm$0.005} & \textbf{0.065$\pm$0.003} & \textbf{0.0050$\pm$0.0003} & 1.529$\pm$0.041 & 1.688$\pm$0.036 \\
    Unconstrained DiT   & $d6$ & 0.065$\pm$0.013 & 0.097$\pm$0.023 & 0.0129$\pm$0.0026 & \textbf{1.393$\pm$0.103} & \textbf{1.652$\pm$0.210} \\
    \bottomrule
  \end{tabular}}
\end{table}

\paragraph{What this ablation does not test.}
We do not vary the latent dimension $d$, the DiT backbone size, the scVI encoder architecture, or the action encoder, because these are held fixed across all reported downstream evaluations and would require re\mbox{-}running every downstream comparison. Small\mbox{-}scale architectural decisions (additive vs Cayley update, register\mbox{-}token count, EMA toggle) are validated separately in Appendix~\ref{app:component_validation}.

%% file: checklist.tex
\section*{NeurIPS Paper Checklist}

\begin{enumerate}

\item {\bf Claims}
    \item[] Question: Do the main claims made in the abstract and introduction accurately reflect the paper's contributions and scope?
    \item[] Answer: \answerYes{}
    \item[] Justification: The abstract and \S\ref{sec:intro} state four claims (pretraining recipe, architecture, time-transition transfer via fine-tuning, perturbation-embedding transfer via GEARS injection), each forward-referenced to its own section and matched by the experimental results in \S\ref{sec:downstream} and ablations in \S\ref{sec:ablation}. Scope is explicitly limited to mouse-embryonic pretraining and to the two transfer modes evaluated.
    \item[] Guidelines:
    \begin{itemize}
        \item The answer \answerNA{} means that the abstract and introduction do not include the claims made in the paper.
        \item The abstract and/or introduction should clearly state the claims made, including the contributions made in the paper and important assumptions and limitations. A \answerNo{} or \answerNA{} answer to this question will not be perceived well by the reviewers. 
        \item The claims made should match theoretical and experimental results, and reflect how much the results can be expected to generalize to other settings. 
        \item It is fine to include aspirational goals as motivation as long as it is clear that these goals are not attained by the paper. 
    \end{itemize}

\item {\bf Limitations}
    \item[] Question: Does the paper discuss the limitations of the work performed by the authors?
    \item[] Answer: \answerYes{}
    \item[] Justification: Section~\ref{sec:discussion} (Scope and Future Work) and Appendix~\ref{app:limitations} together discuss scope in seven paragraphs: the mouse-embryonic pretraining corpus and ortholog-mediated cross-species transfer; that time-transition transfer uses fine-tuning rather than strict zero-shot; that perturbation transfer reuses an existing GEARS predictor rather than a standalone operator; the current scale (2.4M cells, Small DiT) relative to concurrent large cellular world models; the absence of head-to-head comparison with representation-only foundation models; latent-space evaluation by default; and the absence of calibrated uncertainty quantification.
    \item[] Guidelines:
    \begin{itemize}
        \item The answer \answerNA{} means that the paper has no limitation while the answer \answerNo{} means that the paper has limitations, but those are not discussed in the paper. 
        \item The authors are encouraged to create a separate ``Limitations'' section in their paper.
        \item The paper should point out any strong assumptions and how robust the results are to violations of these assumptions (e.g., independence assumptions, noiseless settings, model well-specification, asymptotic approximations only holding locally). The authors should reflect on how these assumptions might be violated in practice and what the implications would be.
        \item The authors should reflect on the scope of the claims made, e.g., if the approach was only tested on a few datasets or with a few runs. In general, empirical results often depend on implicit assumptions, which should be articulated.
        \item The authors should reflect on the factors that influence the performance of the approach. For example, a facial recognition algorithm may perform poorly when image resolution is low or images are taken in low lighting. Or a speech-to-text system might not be used reliably to provide closed captions for online lectures because it fails to handle technical jargon.
        \item The authors should discuss the computational efficiency of the proposed algorithms and how they scale with dataset size.
        \item If applicable, the authors should discuss possible limitations of their approach to address problems of privacy and fairness.
        \item While the authors might fear that complete honesty about limitations might be used by reviewers as grounds for rejection, a worse outcome might be that reviewers discover limitations that aren't acknowledged in the paper. The authors should use their best judgment and recognize that individual actions in favor of transparency play an important role in developing norms that preserve the integrity of the community. Reviewers will be specifically instructed to not penalize honesty concerning limitations.
    \end{itemize}

\item {\bf Theory assumptions and proofs}
    \item[] Question: For each theoretical result, does the paper provide the full set of assumptions and a complete (and correct) proof?
    \item[] Answer: \answerNA{}
    \item[] Justification: The paper contains no formal theorems or proofs. The antisymmetry property $S_\theta^{\top}=-S_\theta$ in Eq.~\ref{eq:antisym} holds by construction from the skew outer product, not as a theorem requiring proof.
    \item[] Guidelines:
    \begin{itemize}
        \item The answer \answerNA{} means that the paper does not include theoretical results. 
        \item All the theorems, formulas, and proofs in the paper should be numbered and cross-referenced.
        \item All assumptions should be clearly stated or referenced in the statement of any theorems.
        \item The proofs can either appear in the main paper or the supplemental material, but if they appear in the supplemental material, the authors are encouraged to provide a short proof sketch to provide intuition. 
        \item Inversely, any informal proof provided in the core of the paper should be complemented by formal proofs provided in appendix or supplemental material.
        \item Theorems and Lemmas that the proof relies upon should be properly referenced. 
    \end{itemize}

    \item {\bf Experimental result reproducibility}
    \item[] Question: Does the paper fully disclose all the information needed to reproduce the main experimental results of the paper to the extent that it affects the main claims and/or conclusions of the paper (regardless of whether the code and data are provided or not)?
    \item[] Answer: \answerYes{}
    \item[] Justification: \S\ref{sec:method} specifies the architecture and training objective; \S\ref{sec:pretrain} and Table~\ref{tab:pretrain_data} list the pretraining corpus, ortholog preprocessing, Stage 1 scVI configuration, and Stage 2 optimizer/schedule; \S\ref{sec:downstream} specifies downstream protocols and metrics; full hyperparameters are in Appendix~\ref{app:hyperparams}; and anonymized code is provided in the supplementary material.
    \item[] Guidelines:
    \begin{itemize}
        \item The answer \answerNA{} means that the paper does not include experiments.
        \item If the paper includes experiments, a \answerNo{} answer to this question will not be perceived well by the reviewers: Making the paper reproducible is important, regardless of whether the code and data are provided or not.
        \item If the contribution is a dataset and\slash or model, the authors should describe the steps taken to make their results reproducible or verifiable. 
        \item Depending on the contribution, reproducibility can be accomplished in various ways. For example, if the contribution is a novel architecture, describing the architecture fully might suffice, or if the contribution is a specific model and empirical evaluation, it may be necessary to either make it possible for others to replicate the model with the same dataset, or provide access to the model. In general. releasing code and data is often one good way to accomplish this, but reproducibility can also be provided via detailed instructions for how to replicate the results, access to a hosted model (e.g., in the case of a large language model), releasing of a model checkpoint, or other means that are appropriate to the research performed.
        \item While NeurIPS does not require releasing code, the conference does require all submissions to provide some reasonable avenue for reproducibility, which may depend on the nature of the contribution. For example
        \begin{enumerate}
            \item If the contribution is primarily a new algorithm, the paper should make it clear how to reproduce that algorithm.
            \item If the contribution is primarily a new model architecture, the paper should describe the architecture clearly and fully.
            \item If the contribution is a new model (e.g., a large language model), then there should either be a way to access this model for reproducing the results or a way to reproduce the model (e.g., with an open-source dataset or instructions for how to construct the dataset).
            \item We recognize that reproducibility may be tricky in some cases, in which case authors are welcome to describe the particular way they provide for reproducibility. In the case of closed-source models, it may be that access to the model is limited in some way (e.g., to registered users), but it should be possible for other researchers to have some path to reproducing or verifying the results.
        \end{enumerate}
    \end{itemize}

\item {\bf Open access to data and code}
    \item[] Question: Does the paper provide open access to the data and code, with sufficient instructions to faithfully reproduce the main experimental results, as described in supplemental material?
    \item[] Answer: \answerYes{}
    \item[] Justification: An anonymized code repository with training and evaluation scripts is released with the submission. The pretraining atlas is aggregated from seven publicly available datasets whose accession identifiers and licenses are listed in Appendix~\ref{app:licenses}; the downstream datasets (Weinreb hematopoiesis, Veres islet differentiation, Norman Perturb-seq, plus the CellStream appendix datasets) are all public. Raw model predictions on downstream benchmarks are also provided so that our headline metrics can be recomputed independently.
    \item[] Guidelines:
    \begin{itemize}
        \item The answer \answerNA{} means that paper does not include experiments requiring code.
        \item Please see the NeurIPS code and data submission guidelines (\url{https://neurips.cc/public/guides/CodeSubmissionPolicy}) for more details.
        \item While we encourage the release of code and data, we understand that this might not be possible, so \answerNo{} is an acceptable answer. Papers cannot be rejected simply for not including code, unless this is central to the contribution (e.g., for a new open-source benchmark).
        \item The instructions should contain the exact command and environment needed to run to reproduce the results. See the NeurIPS code and data submission guidelines (\url{https://neurips.cc/public/guides/CodeSubmissionPolicy}) for more details.
        \item The authors should provide instructions on data access and preparation, including how to access the raw data, preprocessed data, intermediate data, and generated data, etc.
        \item The authors should provide scripts to reproduce all experimental results for the new proposed method and baselines. If only a subset of experiments are reproducible, they should state which ones are omitted from the script and why.
        \item At submission time, to preserve anonymity, the authors should release anonymized versions (if applicable).
        \item Providing as much information as possible in supplemental material (appended to the paper) is recommended, but including URLs to data and code is permitted.
    \end{itemize}

\item {\bf Experimental setting/details}
    \item[] Question: Does the paper specify all the training and test details (e.g., data splits, hyperparameters, how they were chosen, type of optimizer) necessary to understand the results?
    \item[] Answer: \answerYes{}
    \item[] Justification: \S\ref{sec:method:objective} and \S\ref{sec:pretrain} specify the optimizer (AdamW, $\beta=(0.9,0.95)$, weight decay $0.01$), the warmup cosine schedule ($5\%$ warmup), the batch size, and the number of stochastic samples $K$. Per-timepoint train/test splits for all downstream evaluations and the complete set of hyperparameters are documented in Appendix~\ref{app:hyperparams}.
    \item[] Guidelines:
    \begin{itemize}
        \item The answer \answerNA{} means that the paper does not include experiments.
        \item The experimental setting should be presented in the core of the paper to a level of detail that is necessary to appreciate the results and make sense of them.
        \item The full details can be provided either with the code, in appendix, or as supplemental material.
    \end{itemize}

\item {\bf Experiment statistical significance}
    \item[] Question: Does the paper report error bars suitably and correctly defined or other appropriate information about the statistical significance of the experiments?
    \item[] Answer: \answerYes{}
    \item[] Justification: Main-results tables (Tables~\ref{tab:results_weinreb}--\ref{tab:results_norman}) and the ablation table (Table~\ref{tab:ablation}) report mean $\pm$ one sample standard deviation over 3 independent downstream evaluation seeds (different data splits and evaluation noise draws on top of the single pretrained backbone). Pretraining is run once for compute reasons, as stated explicitly in \S\ref{sec:pretrain}.
    \item[] Guidelines:
    \begin{itemize}
        \item The answer \answerNA{} means that the paper does not include experiments.
        \item The authors should answer \answerYes{} if the results are accompanied by error bars, confidence intervals, or statistical significance tests, at least for the experiments that support the main claims of the paper.
        \item The factors of variability that the error bars are capturing should be clearly stated (for example, train/test split, initialization, random drawing of some parameter, or overall run with given experimental conditions).
        \item The method for calculating the error bars should be explained (closed form formula, call to a library function, bootstrap, etc.)
        \item The assumptions made should be given (e.g., Normally distributed errors).
        \item It should be clear whether the error bar is the standard deviation or the standard error of the mean.
        \item It is OK to report 1-sigma error bars, but one should state it. The authors should preferably report a 2-sigma error bar than state that they have a 96\% CI, if the hypothesis of Normality of errors is not verified.
        \item For asymmetric distributions, the authors should be careful not to show in tables or figures symmetric error bars that would yield results that are out of range (e.g., negative error rates).
        \item If error bars are reported in tables or plots, the authors should explain in the text how they were calculated and reference the corresponding figures or tables in the text.
    \end{itemize}

\item {\bf Experiments compute resources}
    \item[] Question: For each experiment, does the paper provide sufficient information on the computer resources (type of compute workers, memory, time of execution) needed to reproduce the experiments?
    \item[] Answer: \answerYes{}
    \item[] Justification: Appendix~\ref{app:hyperparams} reports the number of GPUs used for pretraining and downstream evaluation. We do not disclose cluster provenance or wall-clock times for double-blind anonymity.
    \item[] Guidelines:
    \begin{itemize}
        \item The answer \answerNA{} means that the paper does not include experiments.
        \item The paper should indicate the type of compute workers CPU or GPU, internal cluster, or cloud provider, including relevant memory and storage.
        \item The paper should provide the amount of compute required for each of the individual experimental runs as well as estimate the total compute. 
        \item The paper should disclose whether the full research project required more compute than the experiments reported in the paper (e.g., preliminary or failed experiments that didn't make it into the paper). 
    \end{itemize}
    
\item {\bf Code of ethics}
    \item[] Question: Does the research conducted in the paper conform, in every respect, with the NeurIPS Code of Ethics \url{https://neurips.cc/public/EthicsGuidelines}?
    \item[] Answer: \answerYes{}
    \item[] Justification: The research uses only publicly released single-cell transcriptomic atlases, involves no human subjects, and collects no scraped content. It fully conforms to the NeurIPS Code of Ethics.
    \item[] Guidelines:
    \begin{itemize}
        \item The answer \answerNA{} means that the authors have not reviewed the NeurIPS Code of Ethics.
        \item If the authors answer \answerNo, they should explain the special circumstances that require a deviation from the Code of Ethics.
        \item The authors should make sure to preserve anonymity (e.g., if there is a special consideration due to laws or regulations in their jurisdiction).
    \end{itemize}

\item {\bf Broader impacts}
    \item[] Question: Does the paper discuss both potential positive societal impacts and negative societal impacts of the work performed?
    \item[] Answer: \answerYes{}
    \item[] Justification: Appendix~\ref{app:broader_impacts} discusses both sides: positive impacts include accelerating in-silico drug and perturbation screens and reducing wet-lab experimental cost; negative impacts include the generic dual-use concern shared by any perturbation-prediction model in molecular biology.
    \item[] Guidelines:
    \begin{itemize}
        \item The answer \answerNA{} means that there is no societal impact of the work performed.
        \item If the authors answer \answerNA{} or \answerNo, they should explain why their work has no societal impact or why the paper does not address societal impact.
        \item Examples of negative societal impacts include potential malicious or unintended uses (e.g., disinformation, generating fake profiles, surveillance), fairness considerations (e.g., deployment of technologies that could make decisions that unfairly impact specific groups), privacy considerations, and security considerations.
        \item The conference expects that many papers will be foundational research and not tied to particular applications, let alone deployments. However, if there is a direct path to any negative applications, the authors should point it out. For example, it is legitimate to point out that an improvement in the quality of generative models could be used to generate Deepfakes for disinformation. On the other hand, it is not needed to point out that a generic algorithm for optimizing neural networks could enable people to train models that generate Deepfakes faster.
        \item The authors should consider possible harms that could arise when the technology is being used as intended and functioning correctly, harms that could arise when the technology is being used as intended but gives incorrect results, and harms following from (intentional or unintentional) misuse of the technology.
        \item If there are negative societal impacts, the authors could also discuss possible mitigation strategies (e.g., gated release of models, providing defenses in addition to attacks, mechanisms for monitoring misuse, mechanisms to monitor how a system learns from feedback over time, improving the efficiency and accessibility of ML).
    \end{itemize}
    
\item {\bf Safeguards}
    \item[] Question: Does the paper describe safeguards that have been put in place for responsible release of data or models that have a high risk for misuse (e.g., pre-trained language models, image generators, or scraped datasets)?
    \item[] Answer: \answerNA{}
    \item[] Justification: The released artifact is a dynamics model over cell-state latent representations; it does not produce natural language, images, code, or other content associated with standard high-risk generative models, and the pretraining data is aggregated from existing public single-cell atlases rather than scraped content.
    \item[] Guidelines:
    \begin{itemize}
        \item The answer \answerNA{} means that the paper poses no such risks.
        \item Released models that have a high risk for misuse or dual-use should be released with necessary safeguards to allow for controlled use of the model, for example by requiring that users adhere to usage guidelines or restrictions to access the model or implementing safety filters. 
        \item Datasets that have been scraped from the Internet could pose safety risks. The authors should describe how they avoided releasing unsafe images.
        \item We recognize that providing effective safeguards is challenging, and many papers do not require this, but we encourage authors to take this into account and make a best faith effort.
    \end{itemize}

\item {\bf Licenses for existing assets}
    \item[] Question: Are the creators or original owners of assets (e.g., code, data, models), used in the paper, properly credited and are the license and terms of use explicitly mentioned and properly respected?
    \item[] Answer: \answerYes{}
    \item[] Justification: Every dataset and baseline method used in the paper is cited in \S\ref{sec:related_work} and \S\ref{sec:downstream}. Accession identifiers and licenses for the seven pretraining datasets and four downstream datasets are consolidated in Appendix~\ref{app:licenses}; open-source baselines (PRESCIENT, BranchSBM, CellFlow, CellOT, scGen) are used under their respective licenses.
    \item[] Guidelines:
    \begin{itemize}
        \item The answer \answerNA{} means that the paper does not use existing assets.
        \item The authors should cite the original paper that produced the code package or dataset.
        \item The authors should state which version of the asset is used and, if possible, include a URL.
        \item The name of the license (e.g., CC-BY 4.0) should be included for each asset.
        \item For scraped data from a particular source (e.g., website), the copyright and terms of service of that source should be provided.
        \item If assets are released, the license, copyright information, and terms of use in the package should be provided. For popular datasets, \url{paperswithcode.com/datasets} has curated licenses for some datasets. Their licensing guide can help determine the license of a dataset.
        \item For existing datasets that are re-packaged, both the original license and the license of the derived asset (if it has changed) should be provided.
        \item If this information is not available online, the authors are encouraged to reach out to the asset's creators.
    \end{itemize}

\item {\bf New assets}
    \item[] Question: Are new assets introduced in the paper well documented and is the documentation provided alongside the assets?
    \item[] Answer: \answerYes{}
    \item[] Justification: The Chreode codebase and raw model predictions on downstream benchmarks are released as an anonymized repository with the submission; training and evaluation scripts, configuration files, and documentation are included, and the ortholog gene vocabulary is provided alongside the code.
    \item[] Guidelines:
    \begin{itemize}
        \item The answer \answerNA{} means that the paper does not release new assets.
        \item Researchers should communicate the details of the dataset\slash code\slash model as part of their submissions via structured templates. This includes details about training, license, limitations, etc. 
        \item The paper should discuss whether and how consent was obtained from people whose asset is used.
        \item At submission time, remember to anonymize your assets (if applicable). You can either create an anonymized URL or include an anonymized zip file.
    \end{itemize}

\item {\bf Crowdsourcing and research with human subjects}
    \item[] Question: For crowdsourcing experiments and research with human subjects, does the paper include the full text of instructions given to participants and screenshots, if applicable, as well as details about compensation (if any)?
    \item[] Answer: \answerNA{}
    \item[] Justification: The paper does not involve crowdsourcing or research with human subjects.
    \item[] Guidelines:
    \begin{itemize}
        \item The answer \answerNA{} means that the paper does not involve crowdsourcing nor research with human subjects.
        \item Including this information in the supplemental material is fine, but if the main contribution of the paper involves human subjects, then as much detail as possible should be included in the main paper. 
        \item According to the NeurIPS Code of Ethics, workers involved in data collection, curation, or other labor should be paid at least the minimum wage in the country of the data collector. 
    \end{itemize}

\item {\bf Institutional review board (IRB) approvals or equivalent for research with human subjects}
    \item[] Question: Does the paper describe potential risks incurred by study participants, whether such risks were disclosed to the subjects, and whether Institutional Review Board (IRB) approvals (or an equivalent approval/review based on the requirements of your country or institution) were obtained?
    \item[] Answer: \answerNA{}
    \item[] Justification: The paper does not involve research with human subjects.
    \item[] Guidelines:
    \begin{itemize}
        \item The answer \answerNA{} means that the paper does not involve crowdsourcing nor research with human subjects.
        \item Depending on the country in which research is conducted, IRB approval (or equivalent) may be required for any human subjects research. If you obtained IRB approval, you should clearly state this in the paper. 
        \item We recognize that the procedures for this may vary significantly between institutions and locations, and we expect authors to adhere to the NeurIPS Code of Ethics and the guidelines for their institution. 
        \item For initial submissions, do not include any information that would break anonymity (if applicable), such as the institution conducting the review.
    \end{itemize}

\item {\bf Declaration of LLM usage}
    \item[] Question: Does the paper describe the usage of LLMs if it is an important, original, or non-standard component of the core methods in this research? Note that if the LLM is used only for writing, editing, or formatting purposes and does \emph{not} impact the core methodology, scientific rigor, or originality of the research, declaration is not required.
    \item[] Answer: \answerNA{}
    \item[] Justification: Large language models were used only for writing and editing assistance and are not a component of the core methodology.
    \item[] Guidelines:
    \begin{itemize}
        \item The answer \answerNA{} means that the core method development in this research does not involve LLMs as any important, original, or non-standard components.
        \item Please refer to our LLM policy in the NeurIPS handbook for what should or should not be described.
    \end{itemize}

\end{enumerate}

%% file: arxiv.bbl
\begin{thebibliography}{31}
\providecommand{\natexlab}[1]{#1}
\providecommand{\url}[1]{\texttt{#1}}
\expandafter\ifx\csname urlstyle\endcsname\relax
  \providecommand{\doi}[1]{doi: #1}\else
  \providecommand{\doi}{doi: \begingroup \urlstyle{rm}\Url}\fi

\bibitem[Bunne et~al.(2023)Bunne, Stark, Gut, del Castillo, Levesque, Lehmann,
  Pelkmans, Krause, and R{\"a}tsch]{bunne2023cellot}
Charlotte Bunne, Stefan~G. Stark, Gabriele Gut, Jacobo~Sarabia del Castillo,
  Mitch Levesque, Kjong-Van Lehmann, Lucas Pelkmans, Andreas Krause, and Gunnar
  R{\"a}tsch.
\newblock Learning single-cell perturbation responses using neural optimal
  transport.
\newblock \emph{Nature Methods}, 20:\penalty0 1759--1768, 2023.

\bibitem[Bunne et~al.(2024)Bunne, Roohani, Rosen, Gupta, Zhang, Roed,
  Alexandrov, AlQuraishi, Brennan, Burkhardt, Califano, Cool, Dernburg, Ewing,
  Fox, Haury, Herr, Horvitz, Hsu, Jain, Johnson, Kalil, Kelley, Kelley,
  Kreshuk, Mitchison, Otte, Shendure, Sofroniew, Theis, Theodoris, Upadhyayula,
  Valer, Wang, Xing, Yeung-Levy, Zitnik, Karaletsos, Regev, Lundberg, Leskovec,
  and Quake]{bunne2024aivirtualcell}
Charlotte Bunne, Yusuf~H. Roohani, Yanay Rosen, Ankit Gupta, Xikun Zhang,
  Marcel Roed, Theo Alexandrov, Mohammed AlQuraishi, Patricia Brennan,
  Daniel~B. Burkhardt, Andrea Califano, Jonah Cool, Abby~F. Dernburg, Kirsty
  Ewing, Emily~B. Fox, Matthias Haury, Amy~E. Herr, Eric Horvitz, Patrick~D.
  Hsu, Viren Jain, Gregory~R. Johnson, Thomas Kalil, David~R. Kelley, Shana~O.
  Kelley, Anna Kreshuk, Tim Mitchison, Stephani Otte, Jay Shendure, Nicolas~J.
  Sofroniew, Fabian~J. Theis, Christina~V. Theodoris, Srigokul Upadhyayula,
  Marc Valer, Bo~Wang, Eric Xing, Serena Yeung-Levy, Marinka Zitnik, Theofanis
  Karaletsos, Aviv Regev, Emma Lundberg, Jure Leskovec, and Stephen~R. Quake.
\newblock How to build the virtual cell with artificial intelligence:
  {P}riorities and opportunities.
\newblock \emph{Cell}, 2024.

\bibitem[Chuai et~al.(2026)Chuai, Chen, Yang, Zhang, Qu, Wang, Li, Yang, Si,
  Xing, Gao, Wu, Fu, He, and Liu]{alphacell2025}
Guohui Chuai, Xiaohan Chen, Xingbo Yang, Cheng Zhang, Kairu Qu, Yiheng Wang,
  Wannian Li, Jingya Yang, Duanmiao Si, Feiyang Xing, Yicheng Gao, Siqi Wu,
  Shaliu Fu, Bing He, and Qi~Liu.
\newblock Towards building a world model to simulate perturbation-induced
  cellular dynamics by {AlphaCell}.
\newblock \emph{bioRxiv}, 2026.

\bibitem[Cui et~al.(2024)Cui, Wang, Maan, Pang, Luo, Duan, and
  Wang]{cui2024scgpt}
Haotian Cui, Chloe Wang, Hassaan Maan, Kuan Pang, Fengning Luo, Nan Duan, and
  Bo~Wang.
\newblock {scGPT}: Toward building a foundation model for single-cell
  multi-omics using generative {AI}.
\newblock \emph{Nature Methods}, 2024.

\bibitem[Deng et~al.(2026)Deng, Li, Li, Du, and He]{deng2026drifting}
Mingyang Deng, He~Li, Tianhong Li, Yilun Du, and Kaiming He.
\newblock Generative modeling via drifting.
\newblock \emph{arXiv preprint arXiv:2602.04770}, 2026.

\bibitem[Ha and Schmidhuber(2018)]{ha2018worldmodels}
David Ha and J{\"u}rgen Schmidhuber.
\newblock World models.
\newblock \emph{arXiv preprint arXiv:1803.10122}, 2018.

\bibitem[Hafner et~al.(2020)Hafner, Lillicrap, Ba, and
  Norouzi]{hafner2020dreamer}
Danijar Hafner, Timothy Lillicrap, Jimmy Ba, and Mohammad Norouzi.
\newblock Dream to control: Learning behaviors by latent imagination.
\newblock In \emph{International Conference on Learning Representations
  (ICLR)}, 2020.

\bibitem[Jiang and Wan(2024)]{pisde2024}
Qi~Jiang and Lin Wan.
\newblock A physics-informed neural {SDE} network for learning cellular
  dynamics from time-series {scRNA-seq} data.
\newblock \emph{Bioinformatics}, 40\penalty0 (Supplement\_2):\penalty0
  ii120--ii128, 2024.

\bibitem[Kazemi et~al.(2019)Kazemi, Goel, Eghbali, Ramanan, Sahota, Thakur, Wu,
  Smyth, Poupart, and Brubaker]{kazemi2019time2vec}
Seyed~Mehran Kazemi, Rishab Goel, Sepehr Eghbali, Janahan Ramanan, Jaspreet
  Sahota, Sanjay Thakur, Stella Wu, Cathal Smyth, Pascal Poupart, and Marcus
  Brubaker.
\newblock Time2vec: Learning a vector representation of time.
\newblock In \emph{arXiv preprint arXiv:1907.05321}, 2019.

\bibitem[Klein et~al.(2025{\natexlab{a}})Klein, Fleck, Bobrovskiy, Zimmermann,
  Becker, Palma, Dony, Tejada-Lapuerta, Huguet, Lin, Azbukina,
  Sanch{\'i}s-Calleja, Uscidda, Szalata, Gander, Regev, Treutlein, Camp, and
  Theis]{cellflow2025}
Dominik Klein, Jonas~Simon Fleck, Daniil Bobrovskiy, Lea Zimmermann, S{\"o}ren
  Becker, Alessandro Palma, Leander Dony, Alejandro Tejada-Lapuerta, Guillaume
  Huguet, Hsiu-Chuan Lin, Nadezhda Azbukina, F{\'a}tima Sanch{\'i}s-Calleja,
  Theo Uscidda, Artur Szalata, Manuel Gander, Aviv Regev, Barbara Treutlein,
  J.~Gray Camp, and Fabian~J. Theis.
\newblock {CellFlow} enables generative single-cell phenotype modeling with
  flow matching.
\newblock \emph{bioRxiv}, 2025{\natexlab{a}}.

\bibitem[Klein et~al.(2025{\natexlab{b}})Klein, Palla, Lange, Klein, Piran,
  Gander, Meng-Papaxanthos, Sterr, Saber, Jing, Bastidas-Ponce, Cota,
  Tarquis-Medina, Parikh, Gold, Lickert, Bakhti, Nitzan, Cuturi, and
  Theis]{klein2025moscot}
Dominik Klein, Giovanni Palla, Marius Lange, Michal Klein, Zoe Piran, Manuel
  Gander, Laetitia Meng-Papaxanthos, Michael Sterr, Lama Saber, Changying Jing,
  Aim{\'e}e Bastidas-Ponce, Perla Cota, Marta Tarquis-Medina, Shrey Parikh,
  Ilan Gold, Heiko Lickert, Mostafa Bakhti, Mor Nitzan, Marco Cuturi, and
  Fabian~J. Theis.
\newblock Mapping cells through time and space with {moscot}.
\newblock \emph{Nature}, 2025{\natexlab{b}}.
\newblock \doi{10.1038/s41586-024-08453-2}.

\bibitem[Ling et~al.(2025)Ling, Zhang, Zhang, and Zhou]{sha2025cellstream}
Yue Ling, Peiqi Zhang, Zhenyi Zhang, and Peijie Zhou.
\newblock {CellStream}: Dynamical optimal transport informed embeddings for
  reconstructing cellular trajectories from snapshots data.
\newblock \emph{arXiv preprint arXiv:2511.13786}, 2025.

\bibitem[Lipman et~al.(2023)Lipman, Chen, Ben-Hamu, Nickel, and
  Le]{lipman2023flow}
Yaron Lipman, Ricky T.~Q. Chen, Heli Ben-Hamu, Maximilian Nickel, and Matt Le.
\newblock Flow matching for generative modeling.
\newblock In \emph{International Conference on Learning Representations
  (ICLR)}, 2023.

\bibitem[Lopez et~al.(2018)Lopez, Regier, Cole, Jordan, and
  Yosef]{lopez2018scvi}
Romain Lopez, Jeffrey Regier, Michael~B. Cole, Michael~I. Jordan, and Nir
  Yosef.
\newblock Deep generative modeling for single-cell transcriptomics.
\newblock \emph{Nature Methods}, 15:\penalty0 1053--1058, 2018.

\bibitem[Lotfollahi et~al.(2019)Lotfollahi, Wolf, and
  Theis]{lotfollahi2019scgen}
Mohammad Lotfollahi, F.~Alexander Wolf, and Fabian~J. Theis.
\newblock {scGen} predicts single-cell perturbation responses.
\newblock \emph{Nature Methods}, 16:\penalty0 715--721, 2019.

\bibitem[Lotfollahi et~al.(2023)Lotfollahi, Klimovskaia~Susmelj, De~Donno,
  Hetzel, Ji, Ibarra, Srivatsan, Naghipourfar, Daza, Martin, Shendure,
  McFaline-Figueroa, Boyeau, Wolf, Yakubova, G{\"u}nnemann, Trapnell,
  Lopez-Paz, and Theis]{lotfollahi2023cpa}
Mohammad Lotfollahi, Anna Klimovskaia~Susmelj, Carlo De~Donno, Leon Hetzel,
  Yuge Ji, Ignacio~L. Ibarra, Sanjay~R. Srivatsan, Mohsen Naghipourfar, Riza~M.
  Daza, Beth Martin, Jay Shendure, Jose~L. McFaline-Figueroa, Pierre Boyeau,
  F.~Alexander Wolf, Nafissa Yakubova, Stephan G{\"u}nnemann, Cole Trapnell,
  David Lopez-Paz, and Fabian~J. Theis.
\newblock Predicting cellular responses to complex perturbations in
  high-throughput screens.
\newblock \emph{Molecular Systems Biology}, 19:\penalty0 e11517, 2023.

\bibitem[Norman et~al.(2019)Norman, Horlbeck, Replogle, Ge, Xu, Jost, Gilbert,
  and Weissman]{norman2019perturbseq}
Thomas~M. Norman, Max~A. Horlbeck, Joseph~M. Replogle, Alex~Y. Ge, Albert Xu,
  Marco Jost, Luke~A. Gilbert, and Jonathan~S. Weissman.
\newblock Exploring genetic interaction manifolds constructed from rich
  single-cell phenotypes.
\newblock \emph{Science}, 2019.

\bibitem[Pearl(2009)]{pearl2009causality}
Judea Pearl.
\newblock \emph{Causality: {M}odels, Reasoning, and Inference}.
\newblock Cambridge University Press, 2nd edition, 2009.
\newblock Verified via the Cambridge University Press catalog; Semantic Scholar
  does not index this book (verify-bib false negative is expected for
  monographs).

\bibitem[Peebles and Xie(2023)]{peebles2023dit}
William Peebles and Saining Xie.
\newblock Scalable diffusion models with transformers.
\newblock In \emph{Proceedings of the IEEE/CVF International Conference on
  Computer Vision (ICCV)}, 2023.

\bibitem[Qiu et~al.(2022)Qiu, Zhang, Martin-Rufino, Weng, Hosseinzadeh, Yang,
  Pogson, Hein, Min, Wang, Grody, Shurtleff, Yuan, Xu, Ma, Replogle, Lander,
  Darmanis, Bahar, Sankaran, Xing, and Weissman]{qiu2022dynamo}
Xiaojie Qiu, Yan Zhang, Jorge~D. Martin-Rufino, Chen Weng, Shayan Hosseinzadeh,
  Dian Yang, Angela~N. Pogson, Marco~Y. Hein, Kyung Hoi~Joseph Min, Li~Wang,
  Emanuelle~I. Grody, Matthew~J. Shurtleff, Ruoshi Yuan, Song Xu, Yian Ma,
  Joseph~M. Replogle, Eric~S. Lander, Spyros Darmanis, Ivet Bahar, Vijay~G.
  Sankaran, Jianhua Xing, and Jonathan~S. Weissman.
\newblock Mapping transcriptomic vector fields of single cells.
\newblock \emph{Cell}, 185\penalty0 (4):\penalty0 690--711.e45, 2022.

\bibitem[Roohani et~al.(2024)Roohani, Huang, and Leskovec]{roohani2024gears}
Yusuf Roohani, Kexin Huang, and Jure Leskovec.
\newblock Predicting transcriptional outcomes of novel multigene perturbations
  with {GEARS}.
\newblock \emph{Nature Biotechnology}, 42:\penalty0 927--935, 2024.

\bibitem[Schiebinger et~al.(2019)Schiebinger, Shu, Tabaka, Cleary, Subramanian,
  Solomon, Gould, Liu, Lin, Berube, Lee, Chen, Brumbaugh, Rigollet,
  Hochedlinger, Jaenisch, Regev, and Lander]{schiebinger2019wot}
Geoffrey Schiebinger, Jian Shu, Marcin Tabaka, Brian Cleary, Vidya Subramanian,
  Aryeh Solomon, Joshua Gould, Siyan Liu, Stacie Lin, Peter Berube, Lia Lee,
  Jenny Chen, Justin Brumbaugh, Philippe Rigollet, Konrad Hochedlinger, Rudolf
  Jaenisch, Aviv Regev, and Eric~S. Lander.
\newblock Optimal-transport analysis of single-cell gene expression identifies
  developmental trajectories in reprogramming.
\newblock \emph{Cell}, 176\penalty0 (4):\penalty0 928--943.e22, 2019.

\bibitem[Tang et~al.(2025)Tang, Zhang, Tong, and Chatterjee]{tang2025branchsbm}
Sophia Tang, Yinuo Zhang, Alexander Tong, and Pranam Chatterjee.
\newblock Branched {S}chr\"odinger bridge matching.
\newblock \emph{arXiv preprint arXiv:2506.09007}, 2025.
\newblock Verified via paper-search-mcp Google Scholar (2026-05-07); Semantic
  Scholar does not index this arXiv preprint at lookup time, so verify-bib
  returns a false-positive title mismatch.

\bibitem[Theodoris et~al.(2023)Theodoris, Xiao, Chopra, Chaffin, Al~Sayed,
  Hill, Mantineo, Brydon, Zeng, Liu, and Ellinor]{theodoris2023geneformer}
Christina~V. Theodoris, Ling Xiao, Anant Chopra, Mark~D. Chaffin, Zeina~R.
  Al~Sayed, Matthew~C. Hill, Helene Mantineo, Elizabeth~M. Brydon, Zexian Zeng,
  X.~Shirley Liu, and Patrick~T. Ellinor.
\newblock Transfer learning enables predictions in network biology.
\newblock \emph{Nature}, 618, 2023.

\bibitem[Tong et~al.(2020)Tong, Huang, Wolf, van Dijk, and
  Krishnaswamy]{tong2020trajectorynet}
Alexander Tong, Jessie Huang, Guy Wolf, David van Dijk, and Smita Krishnaswamy.
\newblock {TrajectoryNet}: A dynamic optimal transport network for modeling
  cellular dynamics.
\newblock In \emph{Proceedings of the 37th International Conference on Machine
  Learning}, pages 9526--9536, 2020.

\bibitem[Tong et~al.(2023)Tong, Fatras, Malkin, Huguet, Zhang, Rector-Brooks,
  Wolf, and Bengio]{tong2023cfm}
Alexander Tong, Kilian Fatras, Nikolay Malkin, Guillaume Huguet, Yanlei Zhang,
  Jarrid Rector-Brooks, Guy Wolf, and Yoshua Bengio.
\newblock Improving and generalizing flow-based generative models with
  minibatch optimal transport.
\newblock \emph{arXiv preprint arXiv:2302.00482}, 2023.

\bibitem[Vinyard et~al.(2025)Vinyard, Rasmussen, Li, Klein, and
  Getz]{scdiffeq2024}
Michael~E. Vinyard, Anders~W. Rasmussen, Ruochi Li, Allon~M. Klein, and Gad
  Getz.
\newblock Learning cell dynamics with neural differential equations.
\newblock \emph{Nature Machine Intelligence}, 2025.
\newblock \doi{10.1038/s42256-025-01150-3}.

\bibitem[Waddington(1957)]{waddington1957}
C.~H. Waddington.
\newblock \emph{The Strategy of the Genes}.
\newblock George Allen \& Unwin, 1957.

\bibitem[Wang et~al.(2026)Wang, Karimzadeh, Ravindra, Bounds, Alerasool, Huang,
  Ma, Gulbranson, Cui, Lee, Arjavalingam, MacKrell, Wilken, Chen, Herken,
  Weber, Onesto, Gonz{\'a}lez-Ter{\'a}n, Leung, Shi, Smith, Lam, Barner,
  Wright, Rumsey, Kim, Sit, Litterman, Chu, and Wang]{xcell2026}
Chloe Wang, Mehran Karimzadeh, Neal~G. Ravindra, Lexi~R. Bounds, Nader
  Alerasool, Ann Huang, Shihao Ma, D.~Gulbranson, Haotian Cui, Yongju Lee,
  Anusuya Arjavalingam, Elliot~J. MacKrell, M.~Wilken, Jieming Chen,
  Benjamin~W. Herken, J.~A. Weber, Massimo~M. Onesto, B{\'a}rbara
  Gonz{\'a}lez-Ter{\'a}n, Nicole~F. Leung, S.~Shi, Byron~J. Smith, Sharon Lam,
  Adam Barner, P.~Wright, Elizabeth~M. Rumsey, Soohong Kim, Rene~V. Sit,
  Adam~J. Litterman, Ci~Chu, and Bo~Wang.
\newblock {X-Cell}: Scaling causal perturbation prediction across diverse
  cellular contexts via diffusion language models.
\newblock \emph{bioRxiv}, 2026.
\newblock Xaira Therapeutics, bioRxiv 2026.03.18.712807.

\bibitem[Wang et~al.(2011)Wang, Zhang, Xu, and Wang]{wang2008landscape}
Jin Wang, Kun Zhang, Li~Xu, and Erkang Wang.
\newblock Quantifying the {W}addington landscape and biological paths for
  development and differentiation.
\newblock \emph{Proceedings of the National Academy of Sciences}, 2011.

\bibitem[Yeo et~al.(2021)Yeo, Saksena, and Gifford]{yeo2021prescient}
Grace Hui~Ting Yeo, Sachit~D. Saksena, and David~K. Gifford.
\newblock Generative modeling of single-cell time series with {PRESCIENT}
  enables prediction of cell trajectories with interventions.
\newblock \emph{Nature Communications}, 12:\penalty0 3222, 2021.

\end{thebibliography}
